\documentclass[conference]{IEEEtran}
\usepackage{times}

\usepackage[numbers]{natbib}
\usepackage{multicol}
\usepackage{multirow}
\usepackage[bookmarks=true]{hyperref}

\usepackage{xspace}
\usepackage{xcolor}
\usepackage{graphicx}
\usepackage{graphbox}
\usepackage[export]{adjustbox}
\usepackage{subfig}
\usepackage{booktabs}
\usepackage{makecell}
\usepackage{amssymb}
\usepackage{amsmath}
\usepackage{pifont}

\definecolor{darkgreen}{HTML}{27AE60}
\definecolor{darkred}{HTML}{C0392B}

\usepackage{cleveref}
\crefname{figure}{fig.}{figs.}
\Crefname{figure}{Fig.}{Figs.}
\crefname{table}{tab.}{tabs.}
\Crefname{table}{Tab.}{Tabs.}

\newcommand{\user}{USER\xspace}
\newcommand{\para}[1]{{\vspace{5pt} \bf \noindent #1 \hspace{5pt}}}
\newcommand{\cmark}{\textcolor{darkgreen}{\ding{52}}}
\newcommand{\xmark}{\textcolor{darkred}{\ding{56}}}

\begin{document}

\title{\includegraphics[width=2.0cm]{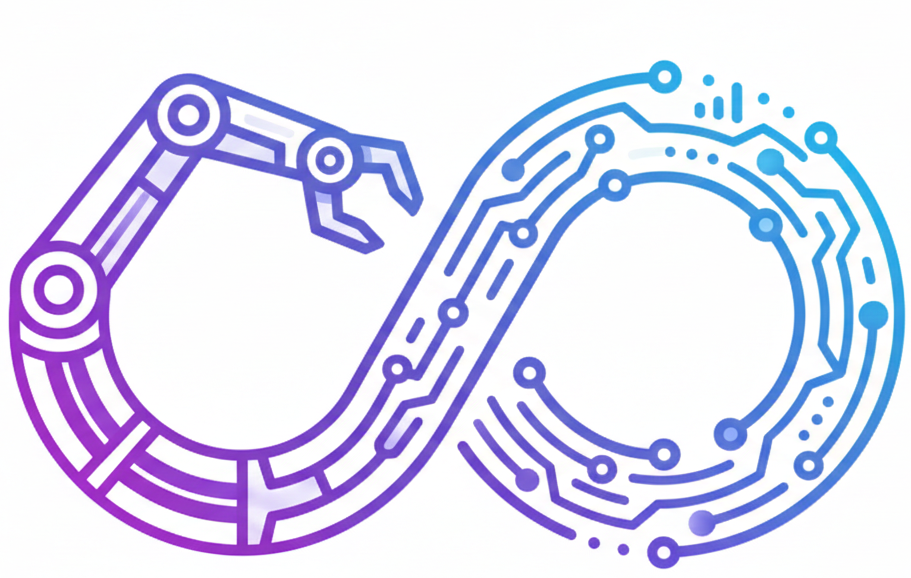} \user: A Unified and Extensible System for Online Real-World Policy Learning in Embodied AI
}

\author{
Hongzhi Zang\textsuperscript{1*}, Shu'ang Yu\textsuperscript{17*}, Hao Lin\textsuperscript{2*}, Tianxing Zhou\textsuperscript{35},  Zefang Huang\textsuperscript{45}, Zhen Guo\textsuperscript{2},\\ Xin Xu\textsuperscript{1}, Yuze Sheng\textsuperscript{1}, Jiakai Zhou\textsuperscript{1}, Si Xu\textsuperscript{2}, Shizhe Zhang\textsuperscript{1}, Feng Gao\textsuperscript{16}, Wenhao Tang\textsuperscript{1}, \\ Yufeng Yue\textsuperscript{3}, Quanlu Zhang\textsuperscript{2}, Xinlei Chen\textsuperscript{1}, Chao Yu\textsuperscript{1\#$\dagger$}, and Yu Wang\textsuperscript{1$\dagger$} \\
\\
\authorblockA{\textsuperscript{1}Tsinghua University \quad \textsuperscript{2}Infinigence AI \quad \textsuperscript{3}Beijing Institute of Technology\\ \textsuperscript{4}Zhejiang University \quad \textsuperscript{5}Zhongguancun Academy \quad 
\textsuperscript{6}Striding.AI \quad \textsuperscript{7}Shanghai AI Laboratory \\
\textsuperscript{*}Equal Contribution \quad \textsuperscript{\#}Project Leader} 
\textsuperscript{$\dagger$}Corresponding Authors: \texttt{yuchao@sz.tsinghua.edu.cn, yu-wang@mail.tsinghua.edu.cn} \\
\textbf{Code}: \url{https://github.com/RLinf/RLinf}.
}

\maketitle

\begin{abstract}
Online policy learning directly in the physical world is a promising yet challenging direction for embodied intelligence. Unlike simulation, real-world systems cannot be arbitrarily accelerated, cheaply reset, or massively replicated, suggesting that real-world policy learning is not merely an algorithmic problem, but inherently a systems problem. We present \user, a \underline{U}nified and extensible \underline{S}yst\underline{E}m for online real-world policy lea\underline{R}ning. On the systems side, \user introduces a hardware abstraction layer for unified robot management and an adaptive communication plane that enables efficient cloud-edge training. On the learning side, \user adopts a fully asynchronous training framework, designs a persistent and cache-aware replay buffer, and provides extensible abstractions for rewards, algorithms, and policies. Experiments in both simulation and the real world demonstrate that \user supports multi-robot coordination, heterogeneous manipulators, cloud-edge training with large models, and long-running asynchronous training. Together, these capabilities establish \user as a unified and extensible systems foundation for online real-world policy learning.

\end{abstract}

\IEEEpeerreviewmaketitle

\section{Introduction}
\label{sec:introduction}

A common paradigm for online policy learning in embodied intelligence is to train policies in simulation and then deploy them in the real world \cite{wagenmaker2024overcoming, tan2018sim}. However, unavoidable gaps in dynamics, sensing, and interaction often cause policies to degrade significantly after transfer \cite{huang2023went, wagenmaker2024overcoming}. This has motivated growing interest in learning policies directly in the physical world. Nevertheless, unlike simulation, real-world online learning cannot be arbitrarily accelerated, reset, or replicated. Robots operate in real time, platforms are heterogeneous, networks are unstable, and experiments are long-running and frequently interrupted \cite{luo2024serl}. These properties make real-world online policy learning not merely an algorithmic challenge, but fundamentally a systems problem that tightly couples physical execution, communication, and optimization.

Several system-level challenges limit scalable real-world learning. First, robots are often treated as external environments, which makes it hard to jointly schedule robots and accelerators in distributed deployments. Second, real-world online learning increasingly relies on cloud–edge infrastructures, particularly for large-scale Vision-Language-Action (VLA) models, with rollout at the edge and training in the cloud. These components span heterogeneous and isolated network domains, causing bandwidth asymmetry and cross-domain latency \cite{dulac2019challenges}. Moreover, the physical world cannot be accelerated, making data efficiency the main bottleneck. Synchronous pipelines are commonly used in simulation and therefore result in poor learning efficiency \cite{bloesch2022towards}. Third, embodied learning is moving toward data-driven training with high-dimensional visual streams and long horizons, which greatly increase data volume and experiment duration. However, existing buffers and pipelines remain memory-centric and short-lived, lacking support for persistence, recovery, and cross-phase data reuse \cite{wang2023gear}.

These challenges call for rethinking real-world robot learning from a systems perspective. Practical physical-world learning should not rely solely on isolated algorithmic advances, but instead require unified abstractions and extensible infrastructure. In this paper, we present \user, a \underline{U}nified and extensible \underline{S}yst\underline{E}m for online \underline{R}eal-world policy learning.

At the system level, \user unifies physical robots, GPUs, and other accelerators under a common hardware abstraction layer, enabling automatic discovery, uniform management, and flexible scheduling of heterogeneous resources for diverse learning workloads. To support large-scale real-world learning across cloud–edge domains, \user further introduces an adaptive communication plane that provides tunneling-based cross-domain connectivity, distributed data channels for traffic localization and bandwidth reduction, and streaming multiprocessor (SM)-budgeted weight synchronization to regulate GPU communication overhead during training.

Beyond system architecture, \user organizes the real-world policy learning as a fully asynchronous learning framework, in which data generation, training, data transmission, and weight synchronization proceed independently rather than through tightly synchronized stages. To support long-horizon and data-intensive learning, \user introduces a persistent and cache-aware buffer that enables streaming trajectory ingestion, recovery, and reuse across training phases. Unlike traditional memory-centric pipelines, this design persistently stores data, enabling long-running experiments to recover from network interruptions and temporary training pauses.
On top of this framework, \user provides extensible abstractions for policies, algorithms, and rewards. Rewards can be specified via rules, human feedback, or learned reward models, while policies range from conventional CNN policies to flow-based generative policies and large VLA models, and support both imitation and reinforcement learning algorithms. Together, \user enables researchers to explore diverse real-world learning paradigms without re-engineering the underlying system infrastructure.

Our contributions are summarized as follows:
\begin{itemize}
    \item A unified and extensible systems framework for real-world online policy learning, featuring a hardware abstraction layer that treats robots as schedulable resources alongside accelerators, together with an adaptive communication plane for stable and efficient cloud--edge deployment across heterogeneous and isolated network domains.
    
    \item A fully asynchronous learning framework with persistent and cache-aware data management, as well as extensible abstractions for rewards, algorithms, and policies, enabling scalable online imitation and reinforcement learning with CNN, generative, and large VLA policies in real-world settings.
    
    \item Extensive simulation and real-world experiments demonstrating support for multi-robot coordination, heterogeneous embodiment, cloud-edge collaboration, and long-running asynchronous training under realistic deployment conditions.
    
    \item An open-source implementation of \user that provides a reusable systems foundation for real-world online policy learning.
\end{itemize}

\section{Related Work}
\label{sec:related}

\subsection{Robot Learning Systems}
\label{subsec:related_sys}

\begin{table}[t]
\centering
\caption{Comparison between \user and other online learning systems.}
\scalebox{1.0}{
    \begin{tabular}{l|ccccc}
    \toprule
     & \makecell[c]{Real \\ robot} & \makecell[c]{Multiple \\ robot} & \makecell{Heterogeneous\\ robot} & \makecell[c]{Large \\ model} & \makecell{Open \\ source} \\
     \midrule
    SERL~\cite{luo2024serl} & \cmark & \xmark & \xmark & \xmark & \cmark \\
    SOP~\cite{pan2026sop} & \cmark & \cmark & \xmark & \cmark & \xmark \\
    Qt-Opt~\cite{kalashnikov2018scalable} & \cmark & \cmark & \xmark & \xmark & \xmark \\
    RLlib~\cite{liang2021rllib} & \xmark & \xmark & \xmark & \xmark & \cmark \\
    TMRL~\cite{tmrl2021} & \xmark & \xmark & \xmark & \xmark & \cmark \\
    \user (ours) & \cmark & \cmark & \cmark & \cmark & \cmark \\
    \bottomrule
    \end{tabular}
}
\label{tab:compare}
\end{table}
The rapid progress of embodied intelligence has been driven by high-parallel simulators \cite{makoviychuk2021isaac, mittal2023orbit, Genesis} and large-scale learning frameworks \cite{moritz2018ray, liang2018rllib, hoffman2020acme, yu2025rlinf}. However, these systems are simulation-centric and rely on synchronous pipelines. In real-world learning, the physical environment cannot be accelerated, making data efficiency the bottleneck and synchronous execution inefficient. To address this, \user adopts a fully asynchronous pipeline that lets robots execute continuously while learning runs in parallel.
Beyond asynchrony, scaling real-world learning requires multi-robot parallelism with proper abstraction and communication. ROS2 \cite{macenski2022robot} and Zenoh \cite{corsaro2023zenoh} offer connectivity but limited learning orchestration, while systems such as SERL \cite{luo2024serl} and Qt-Opt \cite{kalashnikov2018scalable} mainly support single-robot or small-model settings. The most recent related work, SOP~\cite{pan2026sop}, supports large-model training across multiple robots but does not support heterogeneous fleets. \Cref{tab:compare} provides comparisons between \user and existing frameworks.

While many components have appeared in prior frameworks, \user uniquely provides \textit{open-source} support for real-world training on heterogeneous robot fleets with large-model optimization. It natively supports geographically distributed, heterogeneous, cloud-edge training, thereby unlocking real-world RL for emerging robot foundation models.

\subsection{Data Management for Long-Horizon Learning}
\label{subsec:related_data}
High-performance, memory-centric replay buffers, exemplified by Reverb \cite{cassirer2021reverb} and Flashbax \cite{flashbax}, leverage volatile RAM to achieve extreme sampling throughput. However, this reliance on memory constrains their ability to support the management requirements of real-world policy learning systems characterized by long-horizon and massive visual data, particularly with the emergence of VLA models. While recent works such as LeRobot Dataset \cite{cadene2024lerobot} have attempted to facilitate efficient streaming access to static datasets via metadata, the systematic handling of large-scale dynamic data during active policy learning remains unresolved. To address this gap, \user proposes a persistent and cache-aware buffer that maintains high-throughput sampling capabilities while supporting asynchronous disk persistence, enabling arbitrarily large datasets with efficient revisiting of historical data, as well as long-horizon experiments with robust crash recovery in real-world training pipelines.

\begin{figure}[t]
    \centering
    \includegraphics[width=0.95\linewidth]{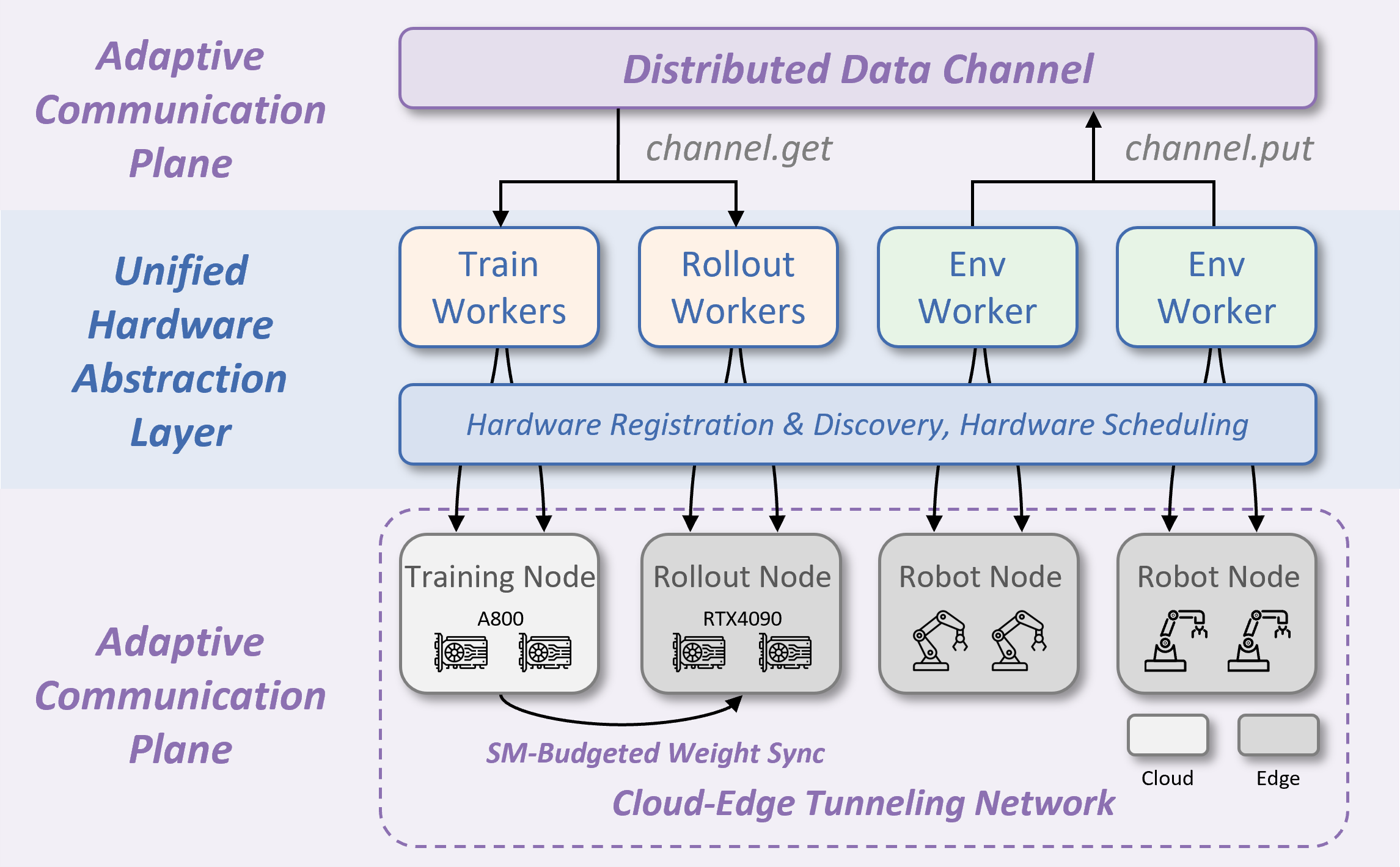}
    \caption{The system architecture design of \user.}
    \label{fig:arch}
\end{figure}

\section{System Architecture Design}
\label{sec:system}

\Cref{fig:arch} illustrates the system architecture of \user, featuring a unified hardware abstraction layer and an adaptive communication plane. Overall, \user abstracts each functional component in online learning as a \textit{worker}, such as the env worker for interaction, the rollout worker for inference, and the actor worker for updating the model. The placement, scheduling, and communication discussed below all operate at the worker level.

\subsection{Unified Hardware Abstraction Layer}
\label{sec:hardware}

\begin{figure}[htbp]
    \centering
    \includegraphics[width=\linewidth]{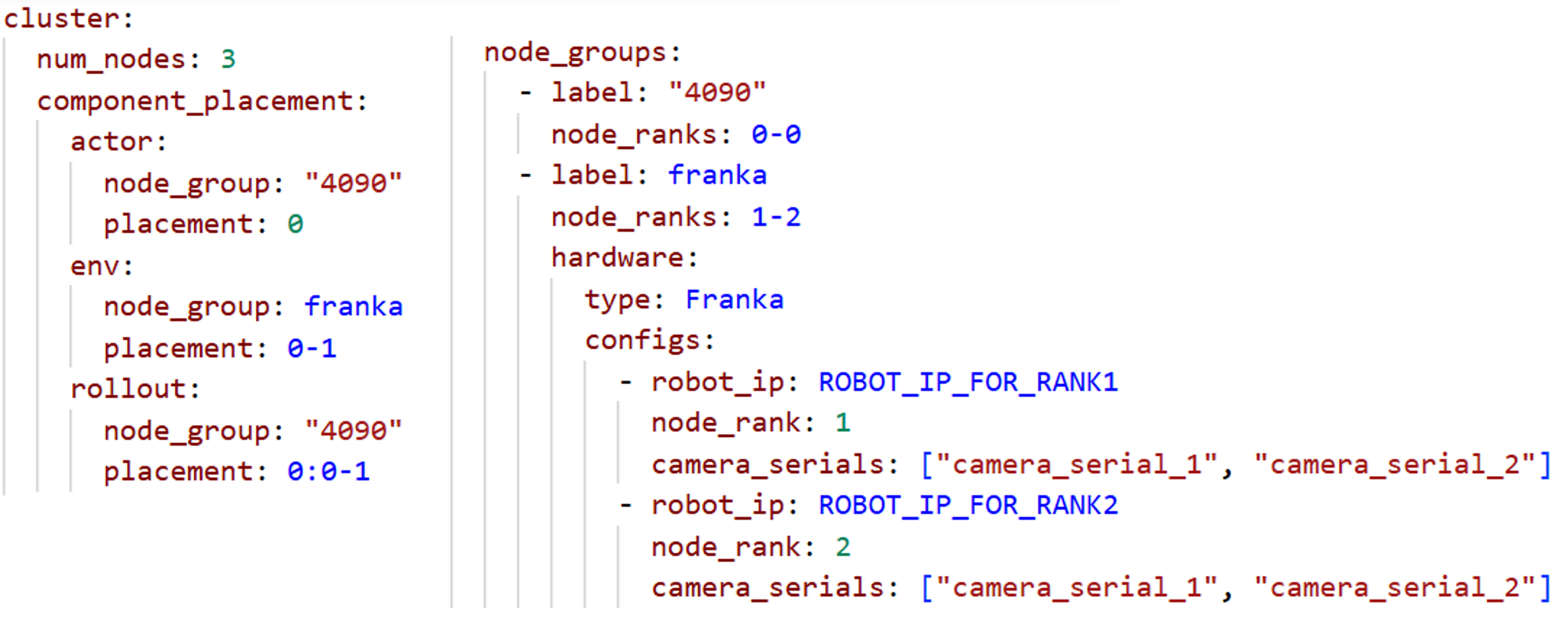}
    \caption{Example robot and accelerator configurations. The system contains two environment nodes, corresponding to two robots, each equipped with two cameras. A single RTX 4090 is shared by both the rollout node and the training node (actor in the figure). }
    \label{fig:placement}
\end{figure}
In pursuit of scalable real-world policy learning, the core idea of \user is to treat physical robots at the same resource level as accelerators (e.g., GPUs and TPUs).
To this end, \user abstracts and manages all types of robots and accelerators uniformly as schedulable hardware units in a special \emph{hardware abstraction layer} (HAL).
The HAL provides unified interfaces for extending new types of hardware, automatically discovering available hardware resources, and scheduling hardware for different learning tasks. 

\para{Node and Hardware Abstraction.}
\user models a deployment as a cluster of nodes, each exporting a set of hardware units. 
A \emph{hardware unit} is the scheduler's atomic allocatable entity---a single GPU device, or a single physical robot, optionally bundled with its required peripherals such as cameras and a spacemouse. 
Each unit is described by a lightweight, typed descriptor (hardware type and model) and configuration metadata such as robot network identity and sensor bindings.
A \emph{node} is inherently heterogeneous, 
    potentially hosting multiple types of hardware units, like different types of robots and GPUs, to maximize flexibility in deployment. In \user, we typically deploy three types of nodes: \textit{rollout nodes} equipped with GPUs for policy inference, \textit{robot nodes} running on CPU-only machines for action execution at the edge, and \textit{training nodes} with large-scale accelerators for centralized training.
Nodes can be organized into \emph{node groups} to capture this heterogeneity,
    where each group contains homogeneous hardware units,
    and a node may belong to multiple groups, depending on its hardware composition.
In this manner, hardware scheduling policies can reason over homogeneous pools while still supporting a heterogeneous cluster. \Cref{fig:placement} shows an example hardware configuration. 

\para{Hardware Registration and Discovery.}
The HAL uses a pluggable \textit{checker} interface to register and extend new hardware. 
Each hardware type supplies a \emph{HAL checker} that defines (i) its type identifier, (ii) how to discover the hardware on a node, and (iii) what metadata is attached to each instance. 
This modularity isolates device-specific logic (GPU vendor tooling, robot connectivity checks, sensor discovery) from the rest of the system. 
At cluster initialization, \user constructs a global hardware inventory by launching a lightweight \emph{hardware probe} process on each node and invoking the registered HAL checkers to discover available hardware.
This discovery can either be \emph{automatic} (typically for PCIe- or USB-connected devices like GPUs, cameras, and spacemouses), or \emph{configuration-driven} (typically for IP-bound robots) due to the need for explicit bindings and safety checks.
For physical robots, discovery includes optional validation like network reachability, presence of required cameras, and basic health checks, ensuring that only usable robots enter the schedulable pool. 
The resulting inventory exposes for each node the available units and their ranks, forming the substrate for subsequent scheduling.

\begin{figure}
    \centering
    \includegraphics[width=0.8\linewidth]{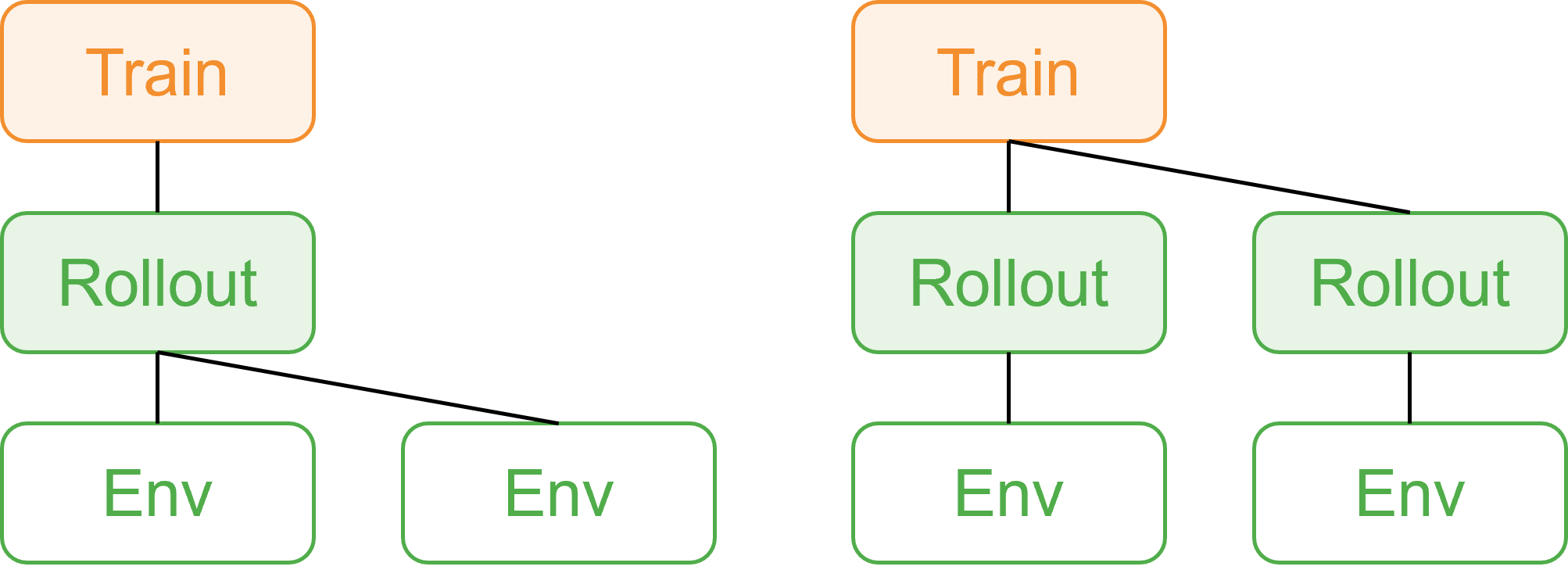}
    \caption{\user supports flexible placement of train, rollout, and environment workers. }
    \label{fig:flexible-placement}
\end{figure}

\para{Hardware Scheduling.}
\user schedules robots and accelerators through a single rank-based placement interface---each component selects node groups and a set of resource ranks, 
where ranks resolve to accelerator or robot units within the node groups' nodes. 
The scheduler then deterministically maps process ranks to these resource ranks (evenly sharing units or assigning multiple units per process) and launches each process with an explicit binding to only its assigned hardware (e.g., constrained visible GPU devices or injected robot endpoint configuration). 
This uniform mechanism enables heterogeneous placements in one job, such as training on one GPU pool while binding different subsets of rollout processes to different types of robots.

This placement flexibility is particularly important in real-world deployments with complex network topologies. For example, in the two-env case shown in \Cref{fig:flexible-placement}, both configurations are feasible: the left configuration reduces weight synchronization overhead, while the right configuration reduces communication overhead between the environment and rollout nodes. As a result, the optimal placement strategy depends on the actual bandwidth and communication characteristics between nodes. \user enables exploring and realizing efficient setups.

\begin{figure}[!h]
    \centering
    \subfloat[Centralized]{
        \includegraphics[width=0.45\linewidth]{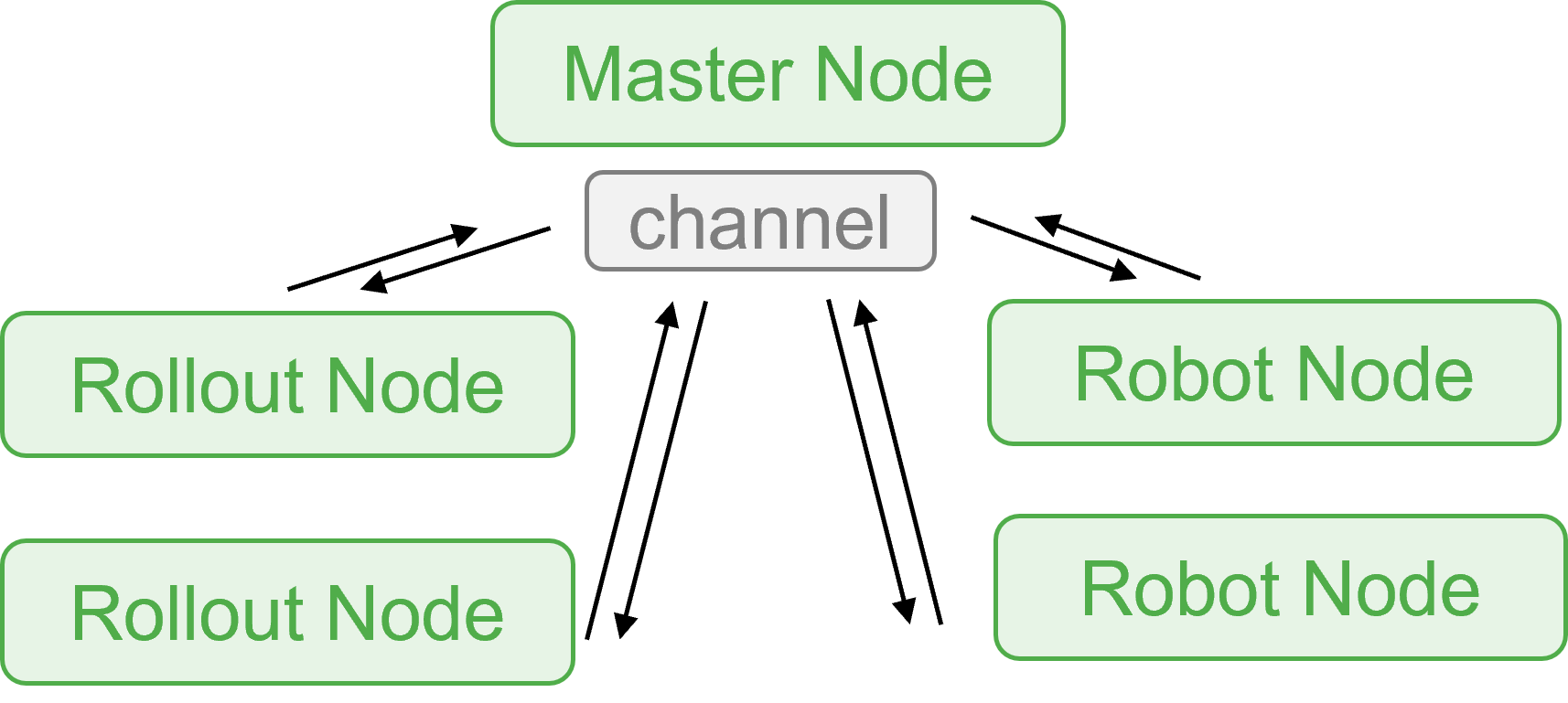}
        \label{fig:centralized-channel}
    }
    \subfloat[Distributed]{
        \includegraphics[width=0.45\linewidth]{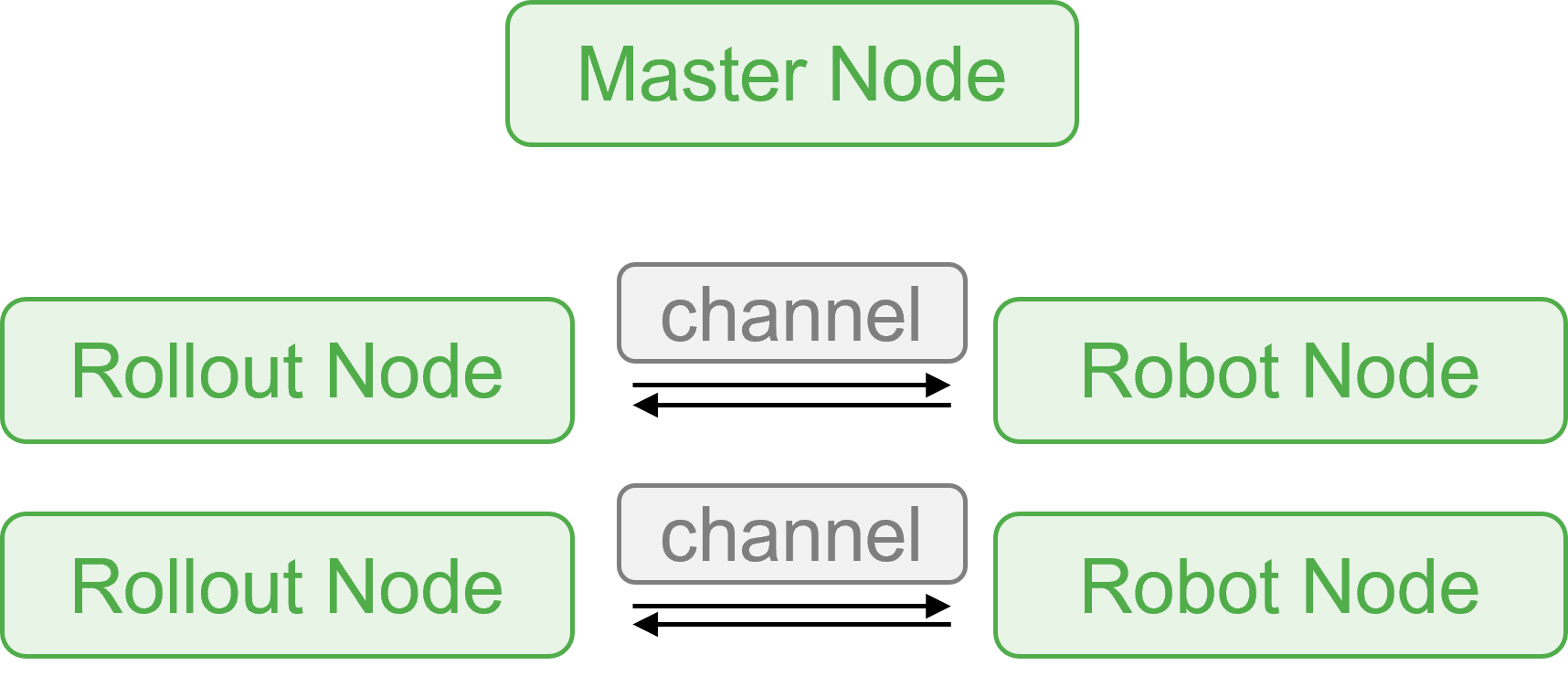}
        \label{fig:dist-channel}
    }
    \caption{Illustration of the centralized and distributed channels.}
    \label{fig:channel}
\end{figure}

\subsection{Adaptive Communication Plane}
\label{sec:comm}

Real-world policy learning runs over cloud–edge distributed compute resources, especially for large-scale VLA models, with rollout and robot nodes at the edge and training nodes in the cloud. \user's \textit{adaptive communication plane} addresses the resulting communication challenges across heterogeneous cloud–edge network domains. \user enables cross-domain connectivity via \emph{tunneling-based cloud–edge networking} for data exchange across network boundaries. To handle uneven bandwidth between intra-domain and cross-domain communication, \user employs a \emph{distributed data channel} to localize traffic and reduce unnecessary cross-domain transfer. In addition, \user introduces \emph{streaming-multiprocessor-budgeted weight synchronization} to prevent NCCL-based communication from monopolizing GPU resources and degrading throughput.

\begin{figure*}[t]
    \centering
    \includegraphics[width=0.98\linewidth]{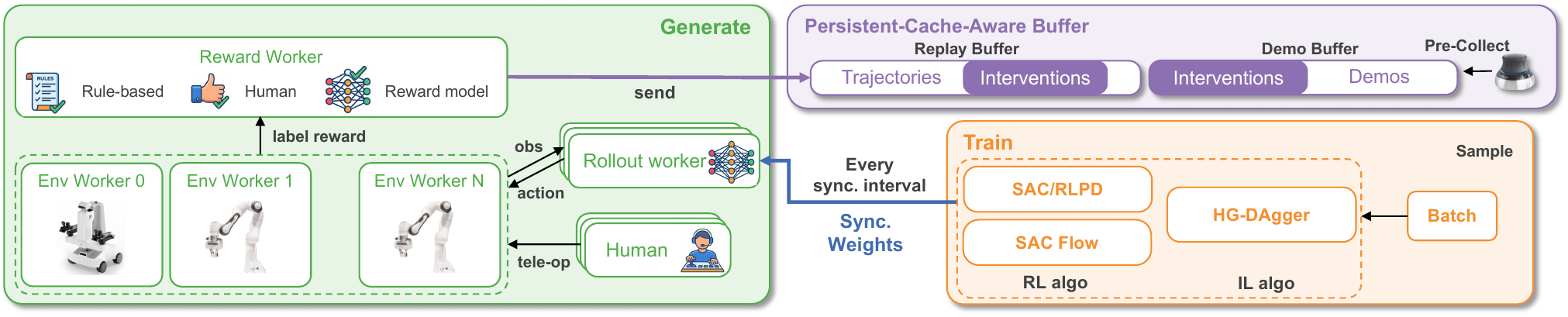}
    \caption{\textbf{Overview of learning framework design:} a fully asynchronous real-world learning pipeline with a persistent, cache-aware buffer and extensible abstractions for policies, algorithms, and reward models. }
    \label{fig:trn-pipeline}
\end{figure*}

\para{Tunneling-based Cloud-Edge Networking.}
\user operates over cloud–edge compute resources across distinct administrative network domains (e.g., NATs, campus networks, factory VLANs) that are inherently isolated and do not support direct communication.
To bridge this gap, \user builds its communication plane on a flattened TCP/IP substrate based on the UDP tunneling technology, implemented with EasyTier~\cite{easytier}.
This design enables all robot, training, and rollout nodes to establish bidirectional connections through the tunnel.
The control plane uses Ray~\cite{moritz2018ray} for cluster membership and worker placement, while the data plane relies on TCP rendezvous to bootstrap point-to-point communication groups; critically, \user binds all control/data traffic to the tunnel interface to make heterogeneous deployments robust to multi-homed hosts and to avoid accidentally routing traffic over slow or firewalled links.

\para{Distributed Data Channel.}
In \user, data exchange among nodes is unified through a \emph{channel} abstraction. A channel represents a dataflow conduit connecting system components and carries observations, actions, states, and intermediate results. Conventional centralized communication (see \Cref{fig:centralized-channel}), where data is first sent to a cloud node and then redistributed to edge nodes, is simple to implement but incurs substantial cross-domain traffic in cloud–edge deployments. Under multi-robot or high-frequency real-world interaction, such designs suffer from high latency and poor stability~\cite{bloesch2022towards, dulac2019challenges}. To address this, \user provides a distributed data \emph{channel} (see \Cref{fig:dist-channel}): a named, first-in-first-out (FIFO) producer–consumer queue hosted by lightweight channel services and accessed via asynchronous put/get APIs. Channels are internally sharded across service instances based on data keys (e.g., robot IDs) to localize traffic within edge or cloud regions and reduce cross-domain transfer. They support multiple producers and consumers, enabling robots and rollout nodes to stream data without direct, synchronous coupling. This design minimizes cross-domain communication and balances load across channel services.

\para{SM-Budgeted Weight Synchronization.}
In \user, when weight synchronization is performed via the NVIDIA Collective Communications Library (NCCL) and transferred directly between GPUs, it occupies GPU execution resources because NCCL’s collectives execute as CUDA kernels that consume streaming multiprocessors (SMs). \user makes this contention explicit and controllable through a tunable configuration that caps the maximum number of NCCL CTAs (cooperative thread arrays) used during weight transfer. By throttling NCCL's SM footprint, \user prevents background weight synchronization from monopolizing GPU execution resources and degrading rollout latency and throughput. This allows \user to sustain stable, low-latency rollouts under asynchronous weight updates.

\section{Learning Framework Design}

Building on the system architecture design that virtualizes and connects heterogeneous robots and compute resources, the learning framework design of \user focuses on how data and computation are organized at runtime. Specifically, it defines a unified framework for real-world learning, including a fully asynchronous pipeline, a persistent and cache-aware buffer, and extensible abstractions for policies, algorithms, and rewards.

\subsection{Fully Asynchronous Pipeline}
\label{sec:async}

In real-world policy learning, the primary bottleneck lies in data collection rather than computation~\cite{luo2024serl}. 
Physical interaction cannot be accelerated, which makes it essential to keep robots executing continuously. 
Synchronous pipelines that tightly couple data generation and training are prone to cascading stalls: delays in training propagate to execution, forcing robots to pause and significantly reducing data efficiency.

\begin{figure}[htp]
    \centering
    \includegraphics[width = 0.98\linewidth]{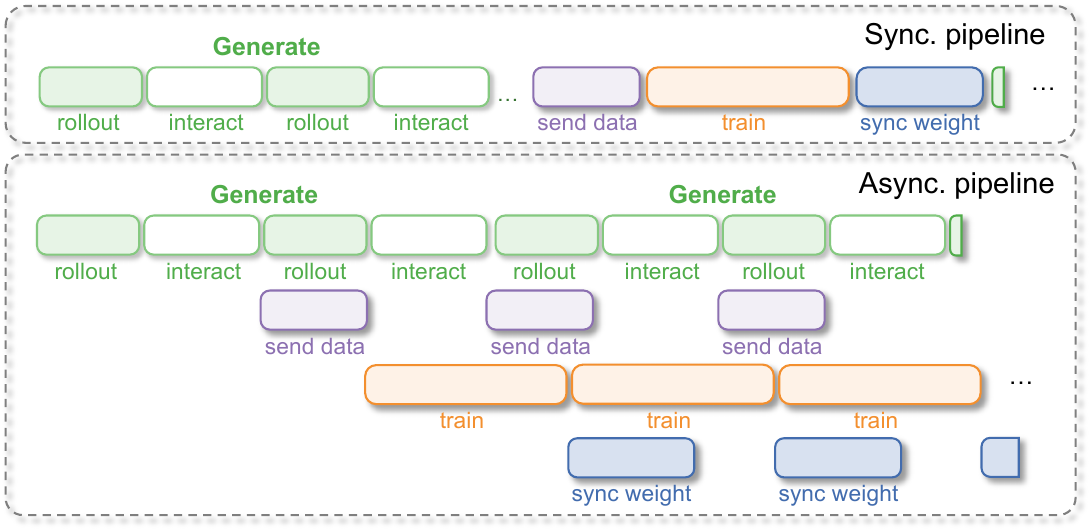}
    \caption{\textbf{Fully Asynchronous pipeline.} \user decouples data generation, training, data transmission, and weight synchronization, significantly improving both data collection and training throughput.}
    \label{fig:async-pipeline}
\end{figure}

As illustrated in \Cref{fig:trn-pipeline} and \Cref{fig:async-pipeline}, \user organizes real-world policy learning as a fully asynchronous pipeline. On the data generation side, multiple environment workers execute policies on physical robots through rollout workers, continuously streaming observations and actions without being blocked by optimization. In parallel, human operators can intervene via teleoperation to provide corrections or demonstrations, while a reward worker assigns supervision signals. All interaction data are asynchronously ingested into a persistent and cache-aware buffer (Sec.~\ref{sec:buffer}), which includes a trajectory replay buffer for autonomous rollouts and a demonstration buffer for pre-collected data and human interventions.

On the training side, learning workers asynchronously sample mini-batches from these buffers to update model parameters, supporting both reinforcement and imitation learning. The updated weights are periodically synchronized back to rollout workers, closing the loop while maintaining uninterrupted robot execution.

\subsection{The Persistent and Cache-Aware Buffer}
\label{sec:buffer}

\begin{figure}[b]
    \centering
    \includegraphics[width=0.75\linewidth]{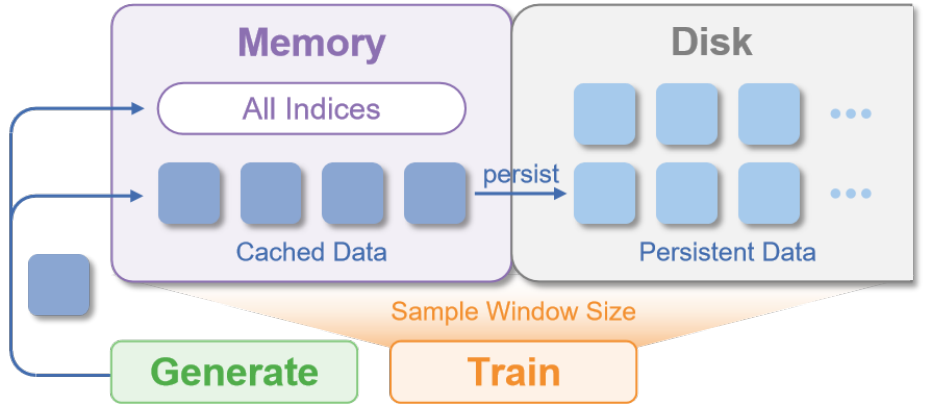}
    \caption{\textbf{The Persistent and Cache-Aware Buffer.} Recent data is stored in memory while historical data is persisted to disk, effectively balancing access efficiency with storage capacity.}
    \label{fig:buffer}
\end{figure}

\begin{figure*}[htp]
  \centering 
  \includegraphics[width = 0.95\linewidth]{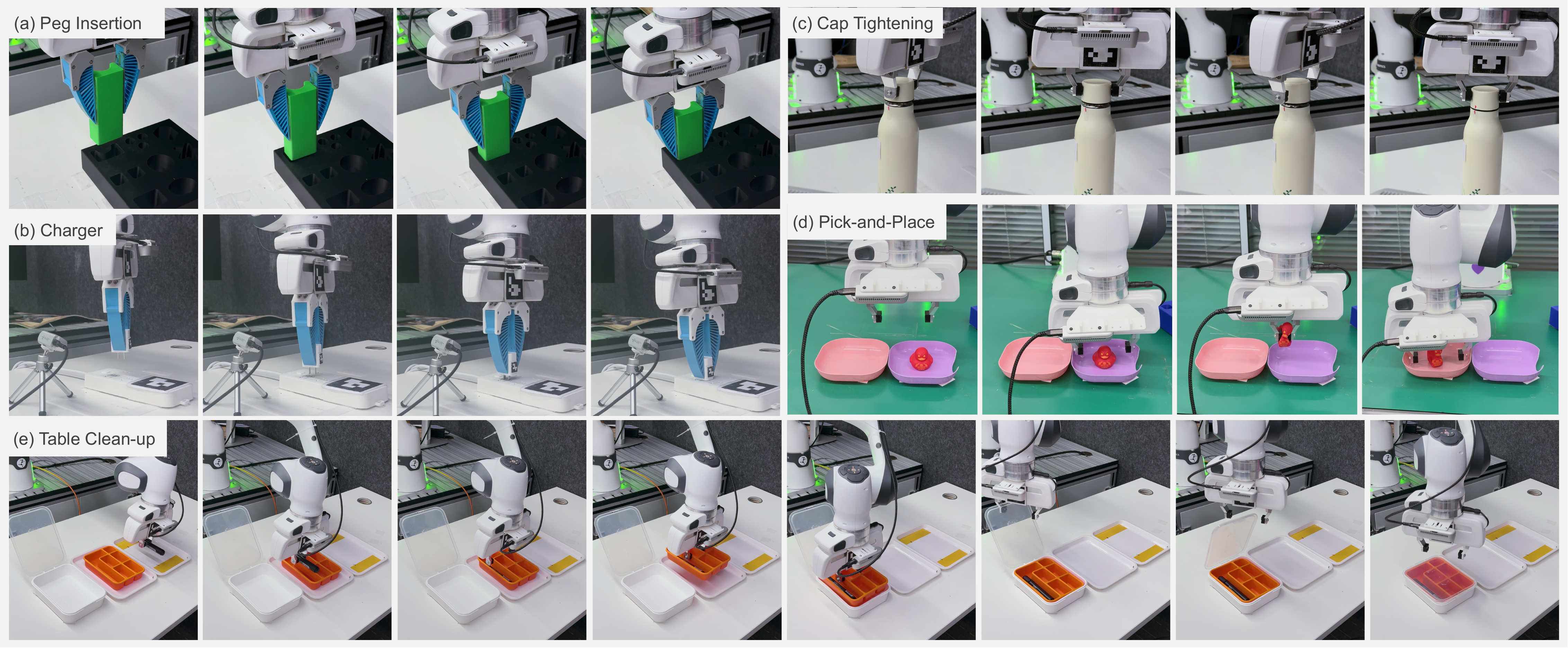}
  \caption{\textbf{Visualization of 5 manipulation tasks.} (a) \textbf{Peg Insertion} inserts a peg into a target hole. (b) \textbf{Charger Plugging} requires contact-rich manipulation with sub-millimeter precision. (c) \textbf{Pick-and-Place} involves grasping and transporting a randomly initialized object (a rubber duck) to a target container. (d) \textbf{Cap Tightening} rotates and tightens a bottle cap to a specified torque or pose. (e) \textbf{Table Clean-up} clears cluttered objects from the tabletop into a designated box, and then closes the lid.}
  \label{fig:tasks}
\end{figure*}

Real-world embodied learning involves long horizons, non-stationary policies, and asynchronous pipelines. Training spans multiple sessions and policy versions, where network failures, restarts, and human intervention are common. Data collected under different policies must often be reused for off-policy updates or recovery, making short-lived, in-memory buffers inadequate for real deployments.

\user adopts a persistent, index-based buffer (\Cref{fig:buffer}) that decouples storage from memory. Trajectories are asynchronously written to disk, while the buffer stores lightweight indices with metadata such as policy version, timestamps, and episode IDs, enabling temporal- and policy-aware sampling over long horizons. To balance efficiency and memory, \user adds a bounded in-memory cache with FIFO replacement. New samples enter the cache first; when full, old entries are evicted but remain indexed on disk. Evicted samples are transparently reloaded when requested, preserving high-throughput sampling with bounded memory.

Unlike in-memory buffers that keep only recent data \cite{cassirer2021reverb, wang2023gear}, the persistent and cache-aware design supports arbitrarily large datasets and retains historical data across evolving policies. Persistence also improves robustness through crash recovery and pipeline decoupling, enabling reliable long-horizon real-world learning beyond volatile memory-centric designs.

\subsection{Extensible Policies, Algorithms, and Rewards}
\label{sec:extensible}
\user is \textit{agnostic} to model architectures and learning algorithms, exposing unified interfaces that enable heterogeneous policies, optimizers, and reward mechanisms to share a single execution and data pipeline.

At the policy level, \user supports both lightweight and large-scale models. Lightweight policies include CNN-based models, such as ResNet-style visual policies \cite{he2016deep}, and expressive flow-matching policies \cite{lipman2022flow} that represent actions via continuous probability flows. At the larger scale, \user integrates VLA models (e.g., $\pi_{0}$/$\pi_{0.5}$ architectures \cite{black2024pi_0,intelligence2025pi_}) that reason over multimodal inputs and output continuous actions. Despite structural and computational differences, all policies are deployed through a unified rollout abstraction.

At the algorithm level, \user supports multiple learning paradigms for robotics. These include off-policy RL such as Soft Actor-Critic (SAC) \cite{haarnoja2018soft}, sample-efficient RL for flow policies (e.g., SAC-Flow \cite{zhang2025sac}), human-in-the-loop methods like RL with pretrained data (RLPD) \cite{ball2023efficient}, and imitation-style updates for large models such as HG-DAgger \cite{kelly2019hg}. Training workers interact with policies through standardized sampling and update APIs, making the optimization layer interchangeable.

Reward specification is also modular. \user supports (i) rule-based task rewards, (ii) human-provided labels, and (iii) learned reward models. Rewards can be attached during rollout or computed offline in post-processing, enabling flexible supervision under real-world constraints.

By unifying policy representations, algorithms, and reward sources, \user enables diverse real-world learning setups—from RL to imitation and human-in-the-loop learning—without re-engineering deployment, data handling, or execution pipelines.

\section{Experiment}
\label{sec:experiment}

\begin{figure*}[t]
    \centering
    \includegraphics[width = 0.3\linewidth]{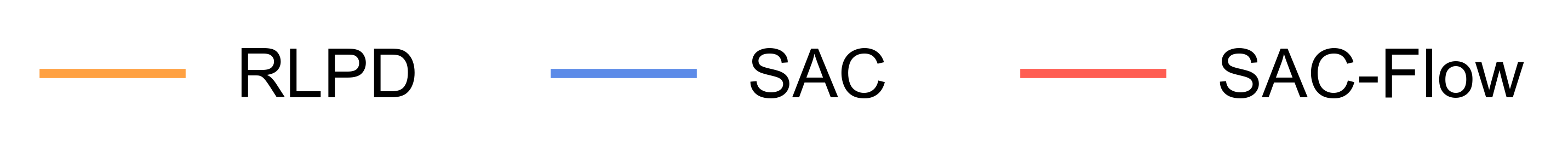}
    \vspace{-5mm}
    \includegraphics[width=0.27\linewidth]{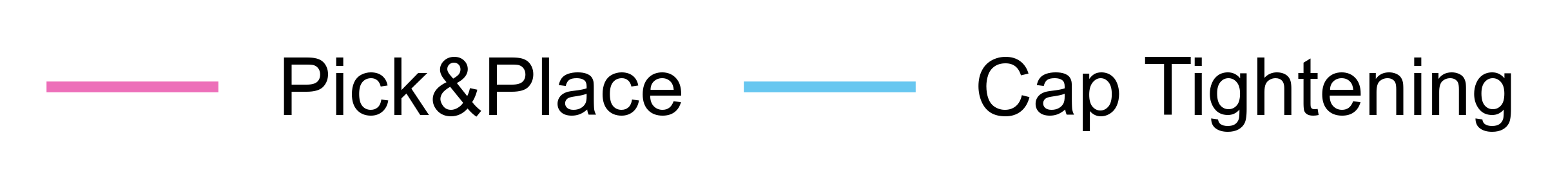}
    \\
    \subfloat[Peg Insertion]{
        \includegraphics[width=0.24\linewidth]{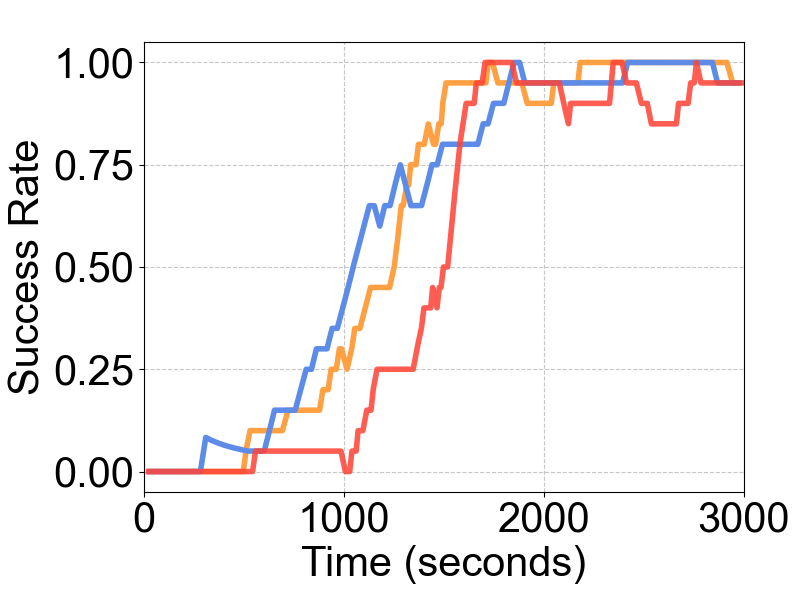}
        \label{fig:peg}
    }
    \subfloat[Charger]{
        \includegraphics[width=0.24\linewidth]{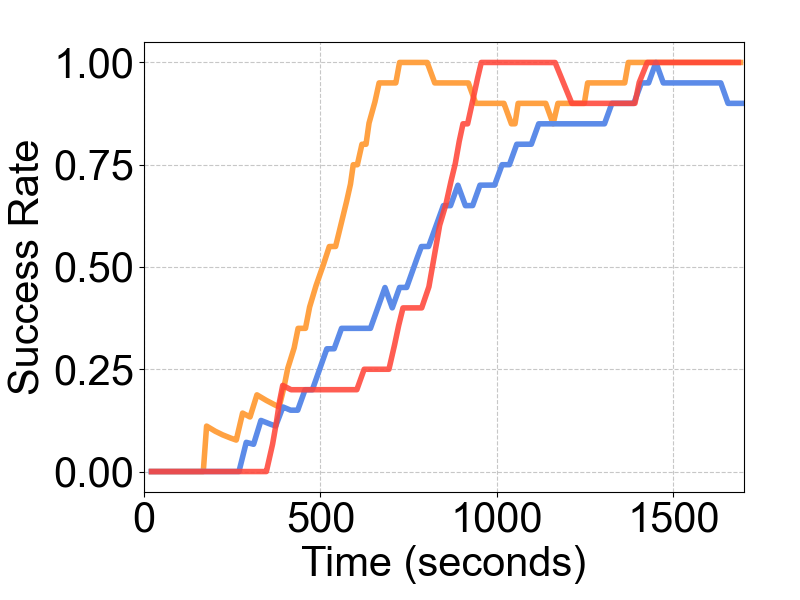}
        \label{fig:charger}
    }
    \subfloat[Other Tasks with RLPD]{
        \includegraphics[width=0.24\linewidth]{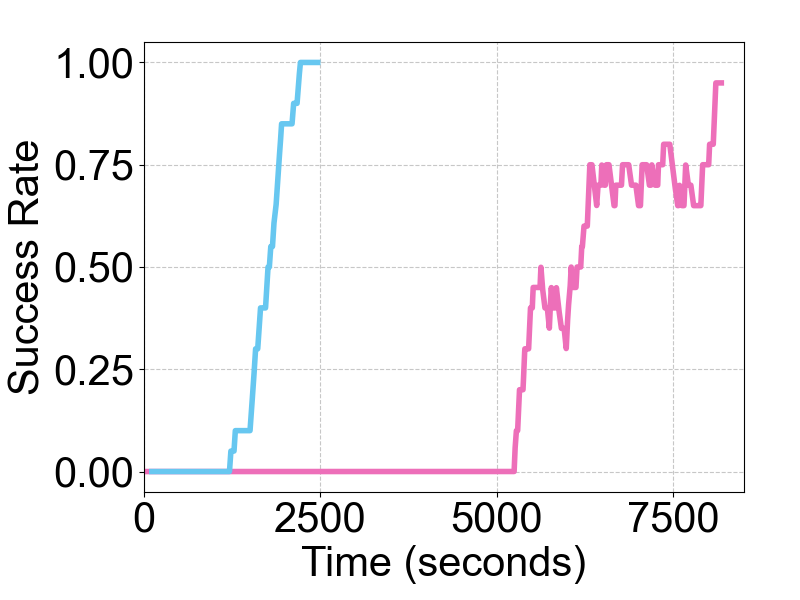}
        \label{fig:more-tasks}
    }

    \caption{Training curves of reinforcement learning on multiple manipulation tasks. The x-axis shows wall-clock time, and the y-axis shows the success rate computed as a moving average with a window size of 20. }
    \label{fig:main-rl}
\end{figure*}

We evaluate \user on a suite of simulated and real-world embodied tasks to validate both its system architecture and learning framework design. Through five groups of experiments, we demonstrate how \user's core components support extensible and reliable real-world online learning:

\begin{enumerate}
\item \textbf{Sec. \ref{exp:extensible}: Extensibility of the learning framework:} \user achieves high performance across multiple tasks while accommodating diverse policies, algorithms, and reward sources within a unified pipeline.
\item \textbf{Sec. \ref{exp:hardware}: Unified hardware abstraction:} \user enables training over multi-robot and heterogeneous robot fleets through its hardware abstraction layer.
\item \textbf{Sec. \ref{exp:comm}: Adaptive communication plane:} \user supports cross-domain edge–cloud collaborative training via its adaptive communication design.
\item \textbf{Sec. \ref{exp:buffer}: Persistent and cache-aware buffer:} \user provides high-capacity and high-throughput storage for long-horizon data ingestion and reuse.
\item \textbf{Sec. \ref{exp:async}: Fully asynchronous pipeline:} \user improves learning efficiency and convergence through its fully asynchronous execution pipeline.
\end{enumerate}

\subsection{Main Results}
\label{exp:extensible}

\begin{table}[hpb]
    \centering
    \caption{Experiment setting summary.}
    \scalebox{0.9}{
    \begin{tabular}{ccccc}
    \toprule
    Task Name & Model & Algorithm & Reward Type & \makecell[c]{Demo\\ (trajs)} \\
    \midrule
    \multirow{3}{*}{Peg Insertion} & \multirow{2}{*}{CNN} & SAC & rule-based, dense & / \\
     &  & RLPD & \multirow{2}{*}{rule-based, sparse} & \multirow{2}{*}{20} \\ 
     & Flow & SAC Flow &  &  \\
     \midrule
    \multirow{3}{*}{Charger} & \multirow{2}{*}{CNN} & SAC & rule-based, dense & / \\
     &  & RLPD & rule-based, sparse & \multirow{2}{*}{20} \\
     & Flow & SAC Flow & rule-based, sparse &  \\
    \midrule
    Cap Tightening & CNN & RLPD & human reward, sparse & 20 \\
    \midrule
    \multirow{2}{*}{Pick-and-Place} & CNN & RLPD & human-reward, sparse & 40 \\
     & $\pi_0$ & HG-DAgger & / & 40 \\
    \midrule
    Table Clean-up & $\pi_0$ & HG-DAgger & / & 40 \\
    \bottomrule
    \end{tabular}
    }
    \label{tab:settings}
\end{table}

\para{Experiment Setup}
We design a suite of five real-world manipulation tasks to evaluate the extensibility of \user’s learning framework, as shown in \Cref{fig:tasks}: \textit{Peg Insertion, Charger, Cap Tightening, Pick-and-Place, and Table Clean-up}. All tasks are conducted on a Franka robotic arm. Detailed task descriptions are provided in Appendix B. 

All experimental settings are summarized in \Cref{tab:settings}.  Experiments with small policies, including CNN and flow-based models, are executed on a local workstation equipped with an RTX 4090 (24GB) GPU, while large VLA policies such as the $\pi_0$ model are full-parameter trained and evaluated on a server with 4 NVIDIA A100 (80GB) GPUs. We select four representative algorithms spanning online reinforcement learning and imitation learning: SAC, RLPD, SAC-Flow, and HG-DAgger. We select the appropriate reward source for each task from rule-based rewards, human-provided rewards, and learned reward models. Rule-based rewards are used for fixed-position manipulation tasks and computed from the end-effector pose. Human rewards are binary, with operators assigning 1 for success and 0 otherwise. The reward model classifies success (1) or failure (0) from observations. Implementation details are in Appendix A.

\begin{figure}[htp]
    \centering
    \includegraphics[width = 0.55\linewidth]{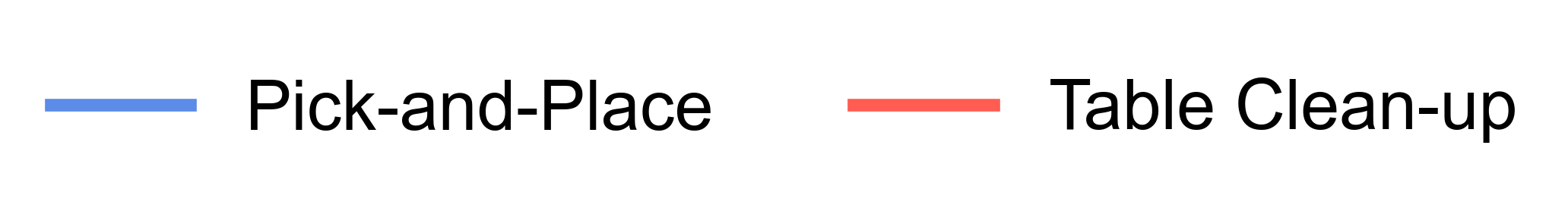}
    \vspace{-5mm}
    \\
    \subfloat[Intervened steps]{
        \includegraphics[width=0.4\linewidth]{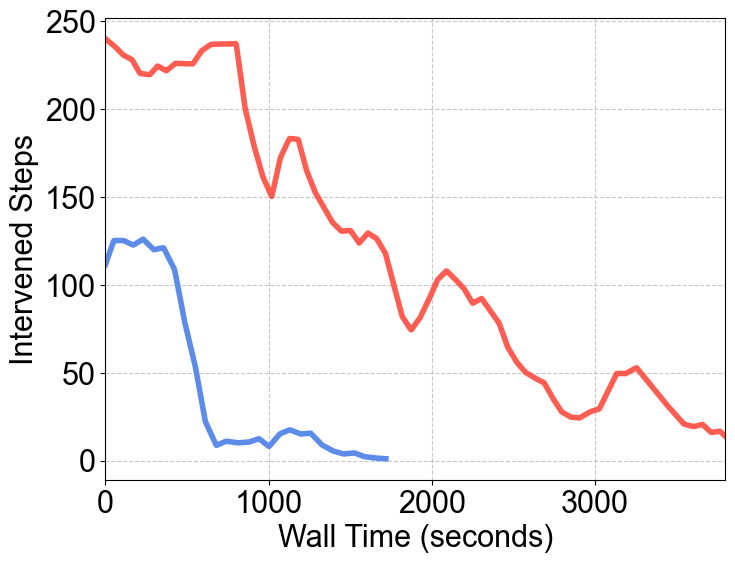}
    }
    \subfloat[Buffer size]{
        \includegraphics[width=0.4\linewidth]{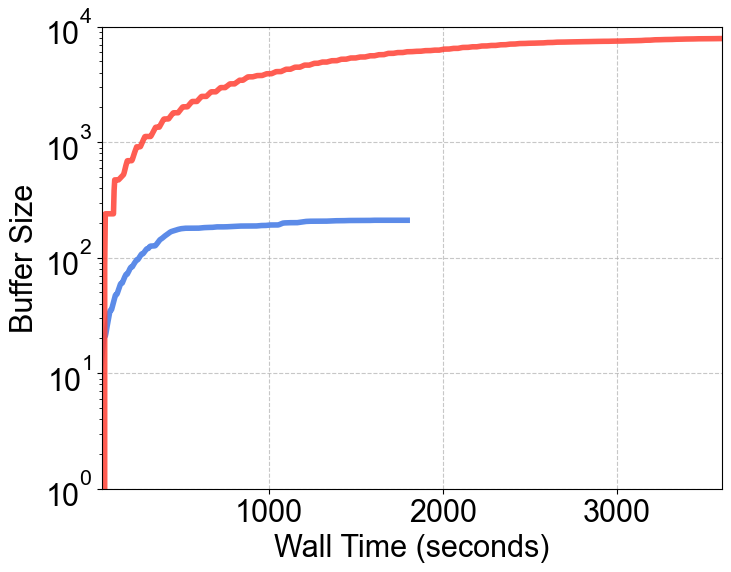}
    }
    \caption{Training curves of imitation learning. A human operator intervenes to guarantee the success of every episode. We report intervened steps per episode as a performance metric. A lower number of intervention steps directly reflects a higher policy performance.}
    \label{fig:main-dagger}
\end{figure}

\para{Task Performance}
\Cref{fig:main-rl} summarizes RL results on multiple manipulation tasks. \Cref{fig:peg,fig:charger} report RLPD, SAC, and SAC-Flow on Peg-Insertion and Charger. Both tasks achieve near-perfect success within 2000s with similar overall performance. SAC performs worse on Charger, likely because higher precision and dense rewards induce suboptimal behaviors.
We further evaluate RLPD on Pick and Place and Cap Tightening (\Cref{fig:more-tasks}). Cap Tightening converges quickly, while Pick and Place involves richer object dynamics and needs longer training. Still, RLPD reaches success rates approaching 1.0 on all tasks.
Unlike RL, imitation learning depends on human demonstrations and intervention. During HG-DAgger training of $\pi_0$, operators ensure success in each episode. As intervention steps approach zero, the policy becomes fully autonomous.
$\pi_0$ is initialized with full-parameter supervised fine-tuning (SFT) and then trained online. \Cref{tab:dagger} compares success before and after training, and \Cref{fig:main-dagger} shows intervention steps and buffer growth. For short-horizon Pick-and-Place, HG-Dagger reaches a 96\% success rate in $\sim$30 minutes using only $\sim$200 online samples.

We compare \user's performance against related real-world learning systems. Results are available in Appendix C.

\begin{table}[htp]
    \centering
    \caption{\user enables significant performance gains after online training for foundation VLA models.}
    \begin{tabular}{ccc}
    \toprule
     & Before online training & After online training \\
    \midrule
    Pick-and-Place & 39/60 & \textbf{58/60} \\
    Table Clean-up & 9/20 & \textbf{16/20} \\
    \bottomrule
    \end{tabular}
    \label{tab:dagger}
\end{table}

\begin{figure}[htp]
    \centering
    \includegraphics[width=0.5\linewidth]{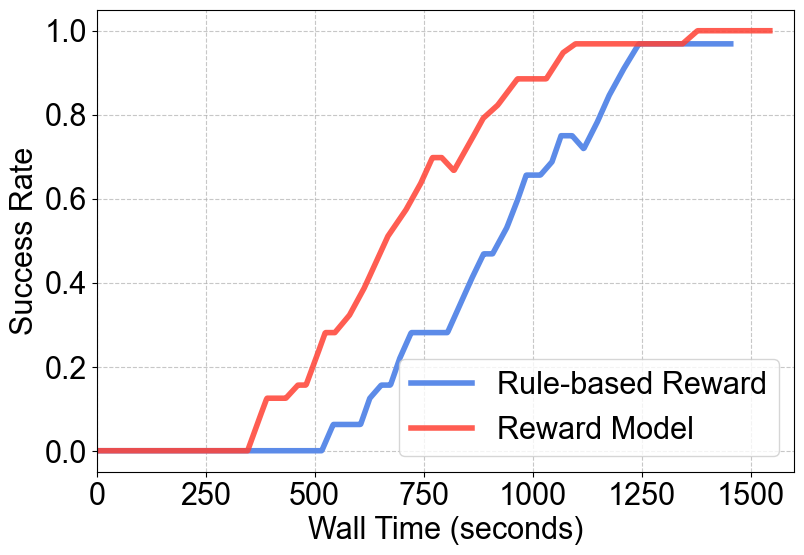}
    \caption{\user supports the reward model to enable automated annotation. In the peg insertion task, the trained reward model provides supervision comparable to human labels.}
    \label{fig:reward-model}
\end{figure}

\para{Reward Model}
To train a reliable reward model, we formulate the reward model as a binary classifier with a pre-trained ResNet18~\cite{he2016deep} backbone. Human operators collect 20 successful trajectories, each remaining stationary for 20 timesteps after task completion to accumulate sufficient positive samples. The final dataset for the reward model comprises approximately 1,600 frames, with a success-to-failure ratio of roughly 1:3. \Cref{fig:reward-model} shows the results of a CNN policy trained with our reward model. Notably, our reward model achieves performance comparable to human rewards on the peg insertion task.

\begin{table*}[htp]
\centering
\caption{Communication performance of distributed channels under cross-domain and same-domain network settings. Enabling distributed channels reduces episode generation time by up to $3\times$ in cross-domain deployments.}
\scalebox{1.0}{
    \begin{tabular}{ccccccccc}
    \toprule
        Domain
        & Distributed 
        & \makecell[c]{Rollout \\ (s/chunk)} 
        & \makecell[c]{Interact \\ (s/chunk)} 
        &  \makecell[c]{Send Obs \\ (s/data)} 
        & \makecell[c]{Send Action \\ (s/data)} &   \makecell[c]{Total Generation Time  $\downarrow$\\ (s/episode) } \\
        \midrule
        \multirow{2}{*}{cross} & w/ &  0.002($\pm$0.000) & 0.106($\pm$0.001) & \textbf{0.042}($\pm$0.031) & \textbf{0.070}($\pm$0.017) & \textbf{21.979}($\pm$0.435) & \\
         & w/o & 0.002($\pm$0.000) & 0.107($\pm$0.001) & 0.671($\pm$0.012) & 0.270($\pm$0.009)& 69.265($\pm$1.905)& \\
         \midrule
        \multirow{2}{*}{same} 

        & w/& 0.006($\pm$0.001)& 0.106($\pm$0.002) & \textbf{0.025}($\pm$0.006)& \textbf{0.028}($\pm$0.008) & \textbf{17.304}($\pm$0.001)\\
        
        & w/o &
        0.007($\pm$0.001)& 0.106($\pm$0.002)& 0.169($\pm$0.018)&
        0.021($\pm$0.008)&
        18.696($\pm$0.710) \\
    \bottomrule
    \end{tabular}
}
\label{tab:dist-chan}
\end{table*}

\subsection{Advantages of a Unified Hardware Layer}
\label{exp:hardware}
With a unified hardware abstraction, \user enables online policy learning on real robots without platform-specific modification. We demonstrate that this design supports efficient multi-robot training and robust learning across heterogeneous embodiments.
\begin{figure}[htp]
    \centering
    \includegraphics[width=0.6\linewidth]{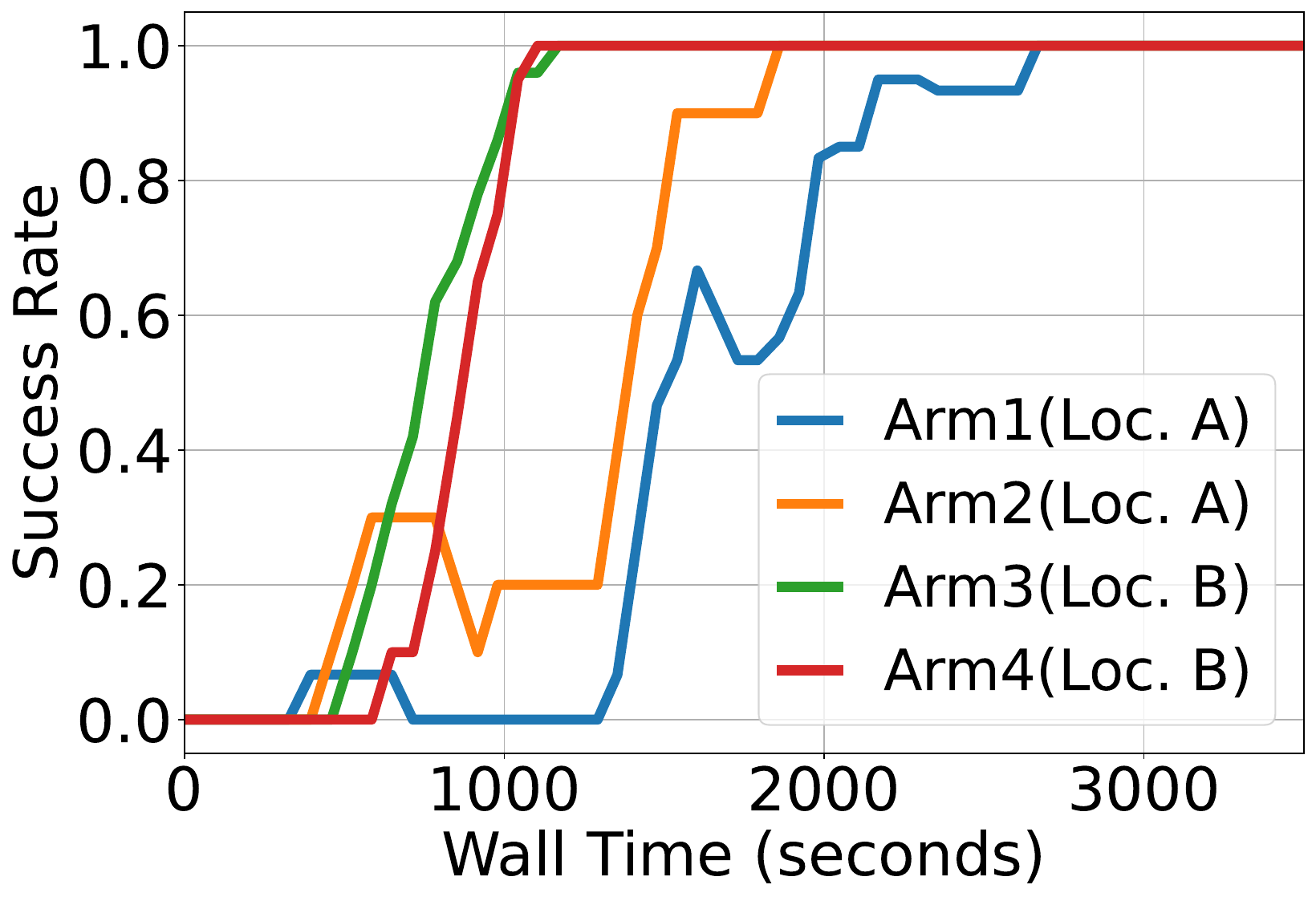}
    \caption{Parallel training on four Franka robot arms. The unified hardware layer enables distributed data collection and learning within a single system framework.}
    \label{fig:multi-robot}
\end{figure}

\para{Training with Multiple Robots}
We first evaluate \user in a multi-robot setting, across two sites (3km apart). Using the unified hardware abstraction, we concurrently train policies on four Franka robot arms executing the peg-insertion task. As shown in \Cref{fig:multi-robot}, all robots achieve 100\% success rate within $\sim$2600 seconds, matching the convergence speed of single-robot baselines. The results indicate that \user can effectively scale real-world training through parallel data collection, improving sample efficiency without degrading learning stability.

\begin{figure}[htp]
    \centering
    \subfloat[Heterogeneous setting]{
        \includegraphics[align=c, width=0.43\linewidth]{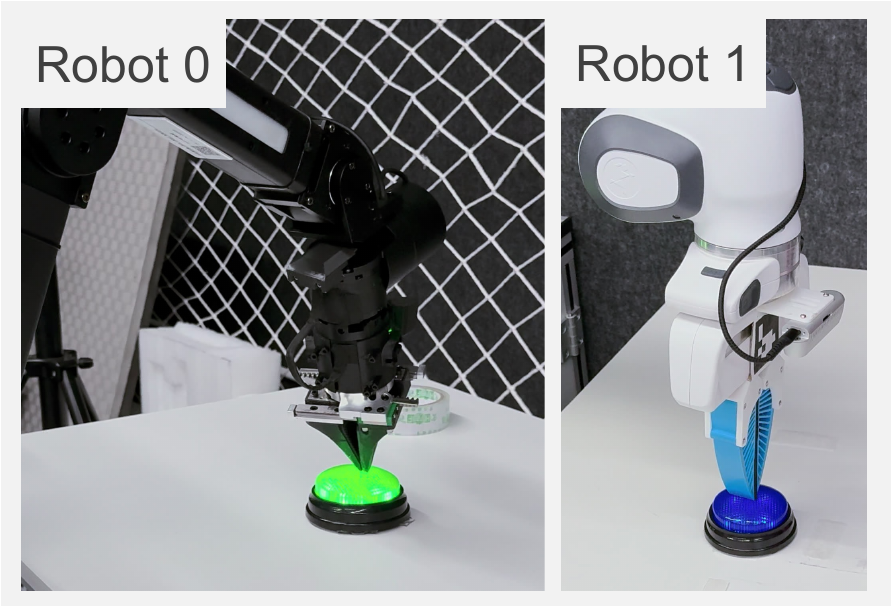}
        \label{fig:hetero-setting}
    }
    \subfloat[Results]{
        \includegraphics[align=c, width=0.4\linewidth]{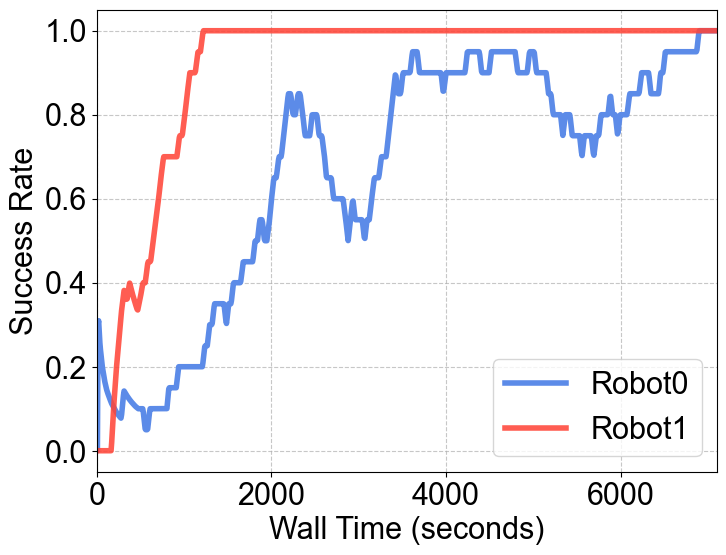}
        \label{fig:hetero-res}
    }
    \caption{Training with two heterogeneous robots. A single unified policy trained on joint data converges successfully, demonstrating cross-embodiment learning.}
    \label{fig:hetero}
\end{figure}

\para{Training with Heterogeneous Robots}
Leveraging heterogeneous platforms enables policies to learn shared visual-semantic representations, thereby improving generalization across embodiments. We use a 7-DoF Franka arm and a 6-DoF low-cost ARX robotic arm to demonstrate the heterogeneous policy capacity. (Hardware setup details in Appendix D). We first unify the observation spaces of the two distinct robot arms as a single image, the end-effector pose, and the normalized gripper state. The action spaces are further unified as the end-effector pose and the normalized gripper action. As shown in \Cref{fig:hetero-setting}, we train a unified CNN-based policy to control two distinct robot arms in a multi-colored button-pressing task. Using SAC with dense rewards, the policy achieves full convergence (\Cref{fig:hetero-res}). Compared to single-robot baselines, heterogeneous training presents significant challenges due to variations in arm DoF, end-effector morphology, camera parameters, and target colors. Consequently, the training process requires approximately two hours to reach convergence.

\subsection{Capability of the Communication Plane}
\label{exp:comm}

In this section, we demonstrate the design of \user's adaptive communication plane through both cross-domain and same-domain experiments.

\begin{table*}[htp]
\centering
\caption{Profiling results of synchronous and asynchronous training pipelines. Generation and training periods denote the intervals between consecutive episode generation and training executions, respectively.}

\scalebox{0.91}{
    \begin{tabular}{ccccccccc}
    \toprule
        &  & \makecell[c]{Rollout \\ (s/chunk)} & \makecell[c]{Interact \\ (s/chunk)} & Send data & \makecell[c]{Train \\ (s/update)} & Sync weights &   \makecell[c]{Generation Period  $\downarrow$\\ (s/episode) } & \makecell{Training Period $\downarrow$\\ (s/update)} \\
        \midrule
        \multirow{3}{*}{\makecell[c]{$\pi_0$ + \\ HG-Dagger}} & Sync & 0.214($\pm$0.001) & 1.093($\pm$0.013) & 0.606($\pm$0.028) & 6.128($\pm$0.247) & 0.816($\pm$0.024) & 45.068($\pm$0.304) & 45.011($\pm$0.251) \\
         & Async & 0.213($\pm$0.016) & 1.091($\pm$0.010) & 7.903($\pm$) & 5.969($\pm$0.196) & 0.789($\pm$0.052) & 37.538($\pm$0.363) & 7.903($\pm$0.234) \\
         & Speed Up & / & / & / & / & / & 1.20$\times$ & 5.70$\times$ \\
        \midrule
        \multirow{3}{*}{\makecell[c]{CNN + \\ SAC}} & Sync & 0.006($\pm$0.002) & 0.108($\pm$0.002) & 0.144($\pm$0.008) & 0.108($\pm$0.002) & 0.162($\pm$0.004) & 20.291($\pm$0.632) & 0.643($\pm$2.984) \\
         & Async & 0.004($\pm$0.004) & 0.107($\pm$0.002) & 0.004($\pm$0.001) & 0.123($\pm$0.005) & 0.174($\pm$0.003) & 13.108($\pm$0.218) & 0.135($\pm$0.034) \\
         & Speed Up & / & / & / & / & / & 1.55$\times$ & 4.61$\times$ \\
        \bottomrule
    \end{tabular}
}
\label{tab:async}
\end{table*}

\para{Cross-Domain Communication}
We evaluate the effectiveness of the distributed data channel under a cross-domain deployment setting. The master node is located in City A and performs policy training, while two nodes in City B handle rollout and robot control, respectively. The two nodes in City B are connected via a high-bandwidth local network, whereas the connection between City A and City B spans thousands of kilometers and incurs high latency and limited bandwidth. This cross-domain deployment reflects a realistic cloud-edge setup, where communication between geographically distributed nodes is enabled via tunneled-based connections.

The first two rows of \Cref{tab:dist-chan} report the cross-domain results. Across both configurations, the rollout and environment interaction times remain similar, as these operations are executed locally without cross-domain communication. In contrast, enabling distributed channels substantially reduces the communication overhead for transmitting observations and actions between the two nodes in City B. As a result, the total time required to generate a single episode is reduced by approximately $3\times$ compared to the centralized baseline.

\para{Same-Domain Communication}
We further evaluate \user in a same-domain setting, where all three nodes are connected via high-speed local networks. The third and fourth rows of \Cref{tab:dist-chan} show the corresponding results. Even under this favorable condition, distributed channels consistently reduce communication overhead. 

Notably, comparing the first and third rows of \Cref{tab:dist-chan} shows that the two nodes in City B under cross-domain deployment achieve communication efficiency comparable to the all-same-domain setup, indicating that \user effectively exploits local high-bandwidth links while avoiding unnecessary cross-domain data transfers.

\begin{figure}[htp]
    \centering
    \subfloat[Throughput v.s. cache ratio]{
        \includegraphics[align=c, width=0.4\linewidth]{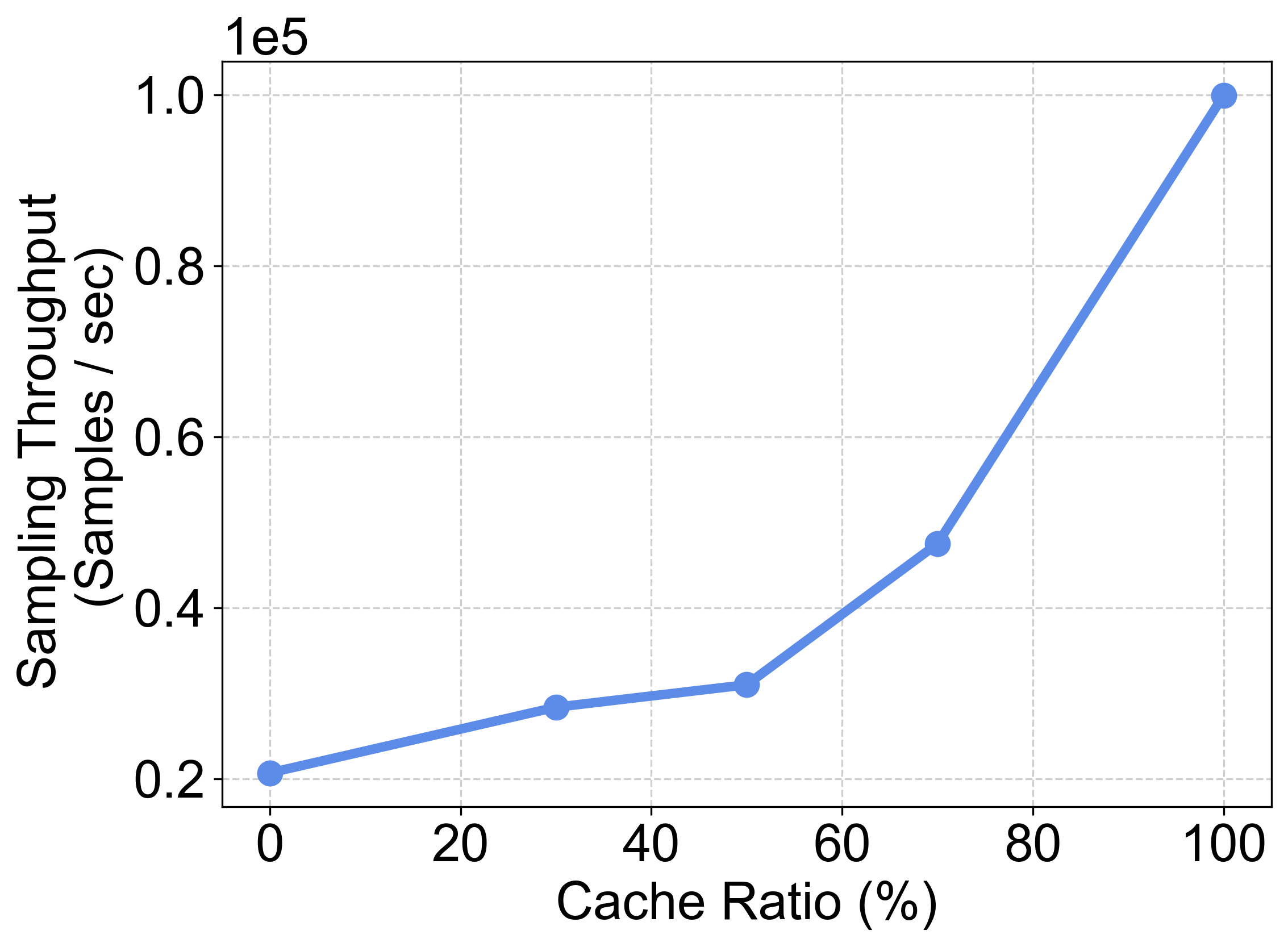}
        \label{fig:throughput-buffer-ratio}
    }
    \subfloat[Throughput v.s. buffer size]{
        \includegraphics[align=c, width=0.58   \linewidth]{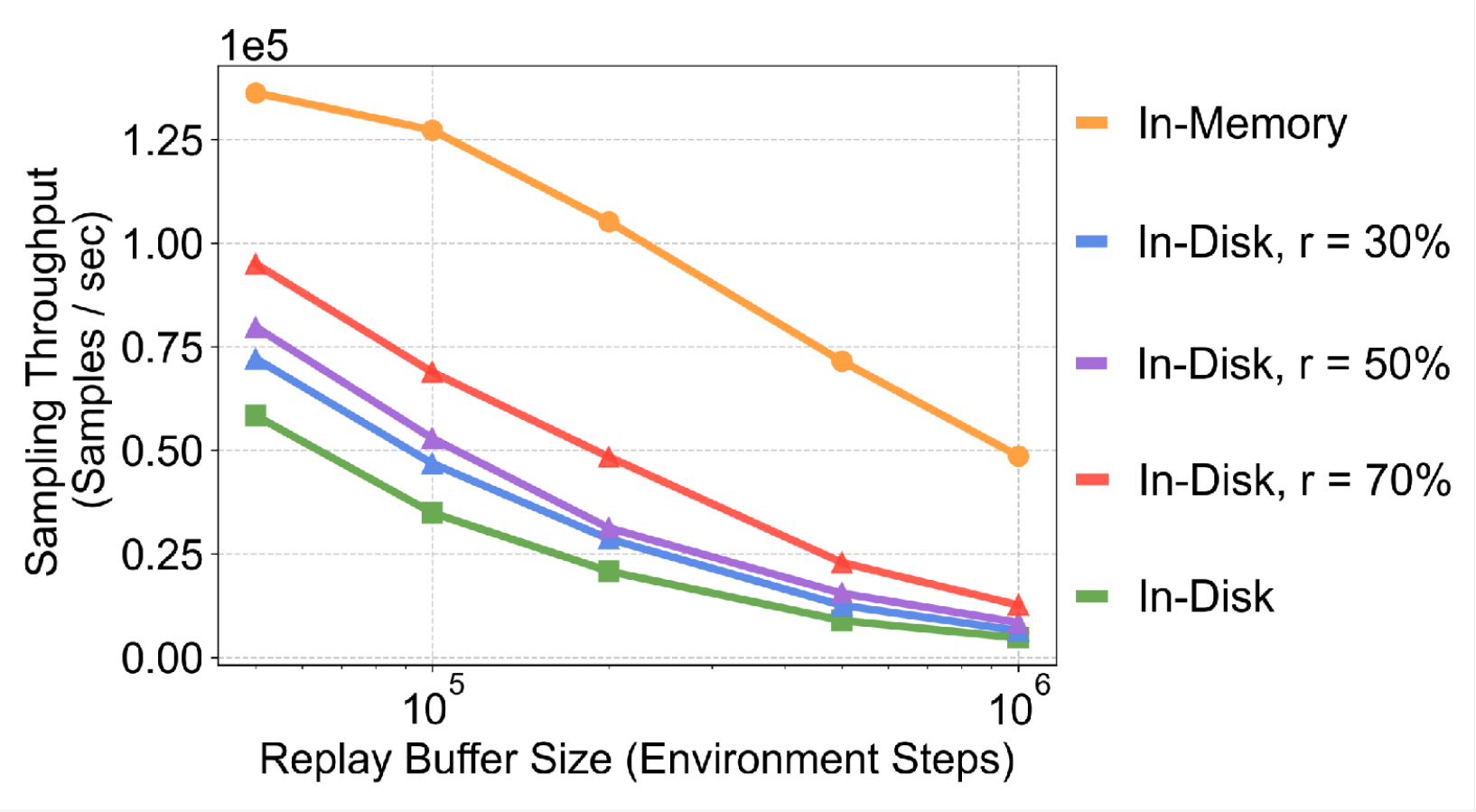}
        \label{fig:throughput-buffer-size}
    }
    \caption{Testing sampling throughput of our persistence and cache-aware buffer.}
    \label{fig:throughput-buffer}
\end{figure}

\subsection{Persistent Training with Cache-aware Buffer}
\label{exp:buffer}

\begin{figure}
    \centering
    \subfloat[Without persistent buffer]{
        \includegraphics[width=0.65\linewidth]{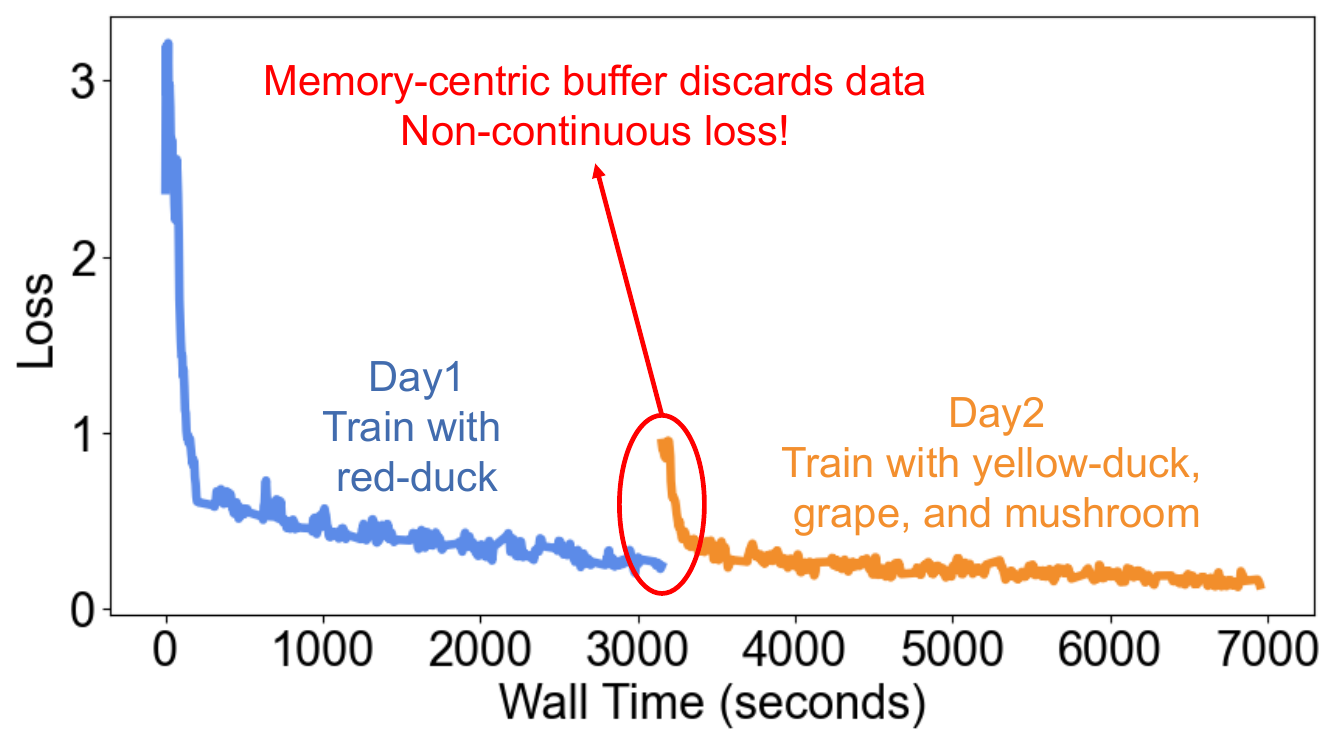}
        \label{fig:without-persistent}
    }
    \\
    \subfloat[With persistent buffer]{
        \includegraphics[width=0.65\linewidth]{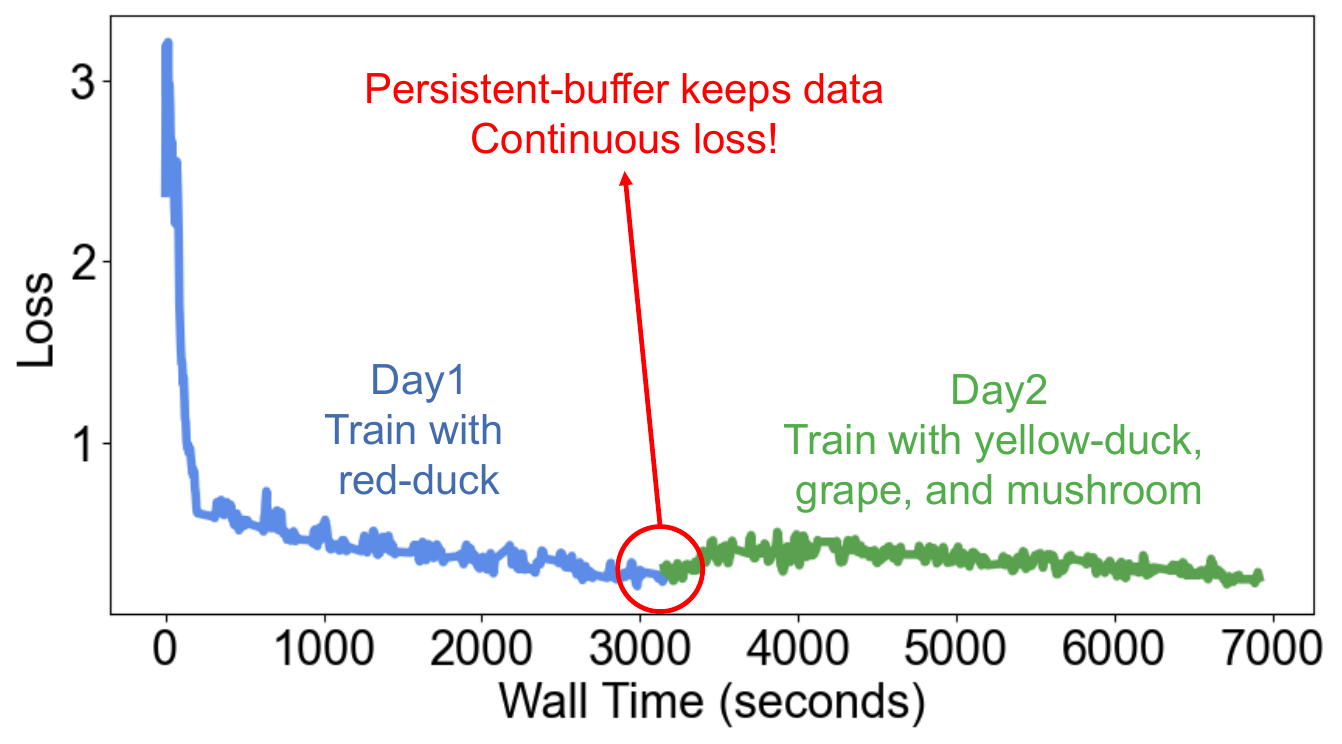}
        \label{fig:with-persistent}
    }
    \caption{DAgger losses during a two-day persistent training with and without the persistent buffer}
    \label{fig:losses-persistent}
\end{figure}
\user's buffer combines persistent storage with an in-memory cache to balance efficiency and capacity. Let $s$ be the buffer size and $c$ the cache size, with ratio $r = c / s$. We profile throughput under different cache ratios in \Cref{fig:throughput-buffer-ratio}, showing that larger $r$ improves throughput. We further evaluate throughput under varying buffer sizes (\Cref{fig:throughput-buffer-size}). A pure in-memory buffer achieves the highest throughput but is memory-limited, while an in-disk buffer provides large capacity at less than half the throughput. Our design strikes a balance, achieving higher throughput than standard disk buffers while retaining large capacity.

This buffer enables continual learning and robust crash recovery. We design a multi-day, multi-task persistent learning task to show the advantage of our buffer. We first initialized the $\pi_0$ model using offline datasets for grasping the red duck to obtain the initial success rate (the first column of \Cref{tab:persistent-training}). On the first day of training, the model interacts with the red duck and improves its performance under human guidance, reaching nearly 100\% success (the second column of \Cref{tab:persistent-training}). On the second day, the model interacted online with three other objects (a yellow duck, a mushroom, and a grape), with and without the persistent buffer. The third and fourth columns of \Cref{tab:persistent-training} report the success rates on different objects after the second-day training under these two settings. The results show that our persistent buffer effectively mitigates forgetting of the object trained on the first day, i.e., the red duck. 

\Cref{fig:losses-persistent} shows the loss curves during continual training. The orange curve, without the persistent buffer, quickly falls below the green curve because the limited non-persistent buffer retains only recent data, making the model more prone to overfitting. \Cref{tab:persistent-training} further confirms that the model forgets how to grasp the red duck. In contrast, the persistent buffer maintains a larger amount of data. Although the green curve has a higher loss than the orange curve, the resulting model achieves better performance across all tasks (\Cref{tab:persistent-training}).

\begin{table}[htbp]
    \centering
    \caption{Persistent buffer enables continual learning without forgetting. Table entries report success rates.}
    \begin{tabular}{l|cccc}
    \toprule
      & \multirow{2}{*}{Initial} & \multirow{2}{*}{After Day 1} & \multicolumn{2}{c}{After Day 2} \\
     ~ & ~ & ~ & w/o persistent & w/ persistent \\
     \midrule
     Red duck & 9/20 & 19/20 & 6/10 & 10/10\\
     Yellow Duck & - & 5/10 & 10/10 & 10/10\\
     Grape & - & 5/10 & 10/10 & 10/10\\
     Mushroom & - & 4/10 & 9/10 & 10/10\\
    \bottomrule
    \end{tabular}
    \label{tab:persistent-training}
\end{table}

\begin{figure}
    \centering
    \subfloat[Convergence speed]{
        \includegraphics[width=0.475\linewidth]{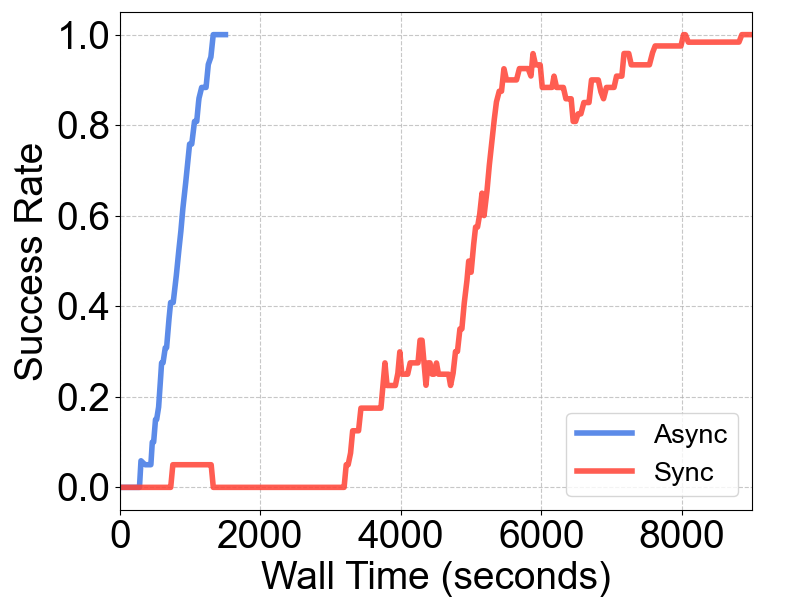}
        \label{fig:async-curve}
    }
    \subfloat[Policy weight sync. intervals]{
        \includegraphics[width=0.475\linewidth]{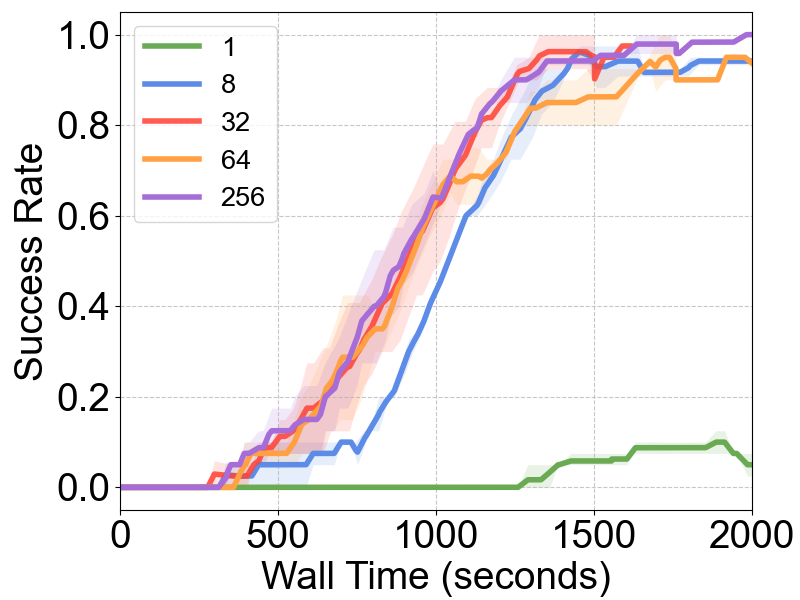}
        \label{fig:abl-syncinterval}
    }
    \caption{Ablation study on asynchronous pipeline design choices for CNN Policy on the Peg-Insertion Task. (a) The async pipeline reduces convergence time from 8,000+ seconds to $\sim$1,500 seconds. (b) Weight synchronization intervals do affect convergence behavior, where interval = 1 is the most unstable setting.}
    \label{fig:async-sync}
\end{figure}

\subsection{Asynchronous Design}
\label{exp:async}

To demonstrate the advantages of our asynchronous framework over traditional synchronous systems, we first profiled the latency of individual stages in both configurations, including rollout, interacting, sending data, training, and syncing weights. As shown in \Cref{tab:async}, although per-stage latencies remain similar, the asynchronous system achieves significantly higher throughput via pipeline overlapping (\Cref{fig:async-pipeline}). Specifically, for $\pi_0$ and CNN models, \user's asynchronous pipeline improves generation throughput by $1.20\times$ and 1.55$\times$, and training throughput by $5.70\times$ and 4.61$\times$, respectively. \Cref{fig:async-curve} further illustrates the disparity in convergence speeds on the peg insertion task, CNN policy, with RLPD. The asynchronous pipeline significantly accelerates training, reducing the convergence time from 8000+ seconds to $\sim$ 1500 seconds.

\para{Ablation for Policy Weights Synchronization Interval}
The weight synchronization interval is a critical design parameter in asynchronous systems. We conducted an ablation study on the peg insertion task using the RLPD algorithm with a CNN policy. As shown in \Cref{fig:abl-syncinterval}, small synchronization intervals (like 1 and 8) cause frequent in-episode weight updates. This induces policy non-stationarity, resulting in slower convergence or complete divergence. Conversely, larger synchronization intervals ensure a stable update.

\section{Conclusion} 
\label{sec:conclusion}

This paper presents \user, a unified and extensible system for real-world online policy learning. \user places robots at the same hardware resource level as accelerators and integrates adaptive communication, persistent cache-aware buffering, and a fully asynchronous training pipeline into a single learning infrastructure. By remaining agnostic to policy architectures and optimization methods, \user supports diverse paradigms—from imitation learning and reinforcement learning to human-in-the-loop learning—deployed on heterogeneous robots within a unified pipeline.
Experiments on both simulated and real-world benchmarks show that \user enables efficient policy learning over heterogeneous multi-robot fleets, provides robust edge–cloud cross-domain communication, and maintains high-capacity, high-throughput buffering for long-horizon training.

\newpage
\section*{Acknowledgments}
This work was supported by the National Natural Science Foundation of China (No.62406159, 62325405), Zhongguancun Academy (Grant No. C20250301), Beijing National Research Center for Information Science, Technology (BNRist) and Beijing Innovation Center for Future Chips.
\bibliographystyle{plainnat}
\bibliography{references}

\iftrue
\clearpage
\appendix

\subsection{Algorithm Implementation Details}
\label{app:algos}
In this section, we first clarify the notation in reinforcement learning, and then provide the implementation details and hyperparameters for the four primary algorithms integrated into USER.
\subsubsection{Problem Setting and Notation.}
We consider a partially observable Markov decision process (POMDP) defined by
$(\mathcal{S}, \mathcal{O}, \mathcal{A}, P, r, \gamma)$, where $s_t \in \mathcal{S}$
denotes the latent environment state, $o_t \in \mathcal{O}$ the observation,
$a_t \in \mathcal{A}$ the action, $P(s_{t+1} \mid s_t, a_t)$ the transition
dynamics, $r(s_t, a_t)$ the reward function, and $\gamma \in (0,1)$ the discount
factor. The policy $\pi_\theta(a_t \mid o_t)$ conditions only on observations.

All off-policy algorithms are trained using a replay buffer
$\mathcal{B}$ that stores transitions of the form
$(o_t, a_t, r_t, o_{t+1})$.
For notational simplicity, we omit the time index when it is clear from context
and use $(o, a, r, o')$ to denote a generic transition sampled from $\mathcal{B}$.

We denote the critic network by $Q_\psi(o, a)$, the policy by $\pi_\theta(a \mid o)$,
and the corresponding target networks by $Q_{\bar{\psi}}$.

\begin{table}[htbp]
\centering
\caption{Hyperparameters of SAC}
\label{tab:sac_params}
\begin{tabular}{ll}
\toprule
Parameter & Value \\
\midrule
Critic Network & ResNet10 encoder (shared) + MLP \\
Actor Network & ResNet10 encoder (shared) + MLP \\
Buffer size & 20000 \\
Train start step & 200 \\
Discount Factor ($\gamma$) & 0.96 \\
Batch Size & 256 \\
Critic learning rate & 3e-4 \\
Actor learning rate & 3e-4 \\
$\alpha$ learning rate & 3e-4 \\
Target Entropy & -3 \\
Temperature ($\alpha$) & Auto tune (init 0.01) \\
Target update ratio $\tau$ & 0.005 \\
Target update frequency & 1 \\
Actor update frequency & 4 \\
Weight sync. frequency & 32 \\
\bottomrule
\end{tabular}
\end{table}

\subsubsection{SAC}

Soft Actor-Critic (SAC) is an off-policy actor-critic algorithm based on the
maximum entropy reinforcement learning framework. It optimizes a stochastic
policy by maximizing both the expected return and the policy entropy:
\begin{equation}
J(\pi) = \sum_{t=0}^{T} \mathbb{E}_{(o_t, a_t) \sim \rho_\pi}
\big[ r(o_t, a_t) + \alpha \mathcal{H}(\pi(\cdot \mid o_t)) \big],
\end{equation}
where $\alpha$ is the temperature parameter controlling the strength of entropy $\mathcal{H}$ regularization.

\para{Critic Update.}
The soft Q-function is trained by minimizing the Bellman residual:
\begin{equation}
\begin{aligned}
&L_Q(\psi) =
\mathbb{E}_{(o,a,r,o') \sim \mathcal{B}} \Big[
\big(
Q_\psi(o,a)
\\
-& (r + \gamma \mathbb{E}_{a' \sim \pi_\theta(\cdot \mid o')}
[Q_{\bar{\psi}}(o', a') - \alpha \log \pi_\theta(a' \mid o')])
\big)^2
\Big],
\end{aligned}
\end{equation}
where $a'$ denotes an action sampled from the current policy at the next
observation $o'$, and $Q_{\bar{\psi}}$ is the target critic updated via
exponential moving average(EMA).

\para{Actor Update.}
The policy is updated by minimizing
\begin{equation}
L_\pi(\theta) =
\mathbb{E}_{o \sim \mathcal{B},\, a \sim \pi_\theta(\cdot \mid o)}
\big[
\alpha \log \pi_\theta(a \mid o) - Q_\psi(o, a)
\big].
\end{equation}

Due to the sample efficiency of off-policy learning, SAC has been widely adopted for RL in the real world. The specific hyperparameters used for our implementation, including the network architecture and entropy regularization settings, are summarized in \Cref{tab:sac_params}.

\begin{table}[htbp]
\centering
\caption{Hyperparameters of SAC-Flow}
\label{tab:sacflow_params}
\begin{tabular}{ll}
\toprule
Parameter & Value \\
\midrule
Critic Network & ResNet10 encoder (shared) + MLP \\
Actor Network & ResNet10 encoder (shared) + Flow-T \\
Denoising Steps ($N$) & 4 \\
Decoder dimension & 256 \\
Attention head & 4 \\
Decoder layer & 2 \\
Log std range & [-5, 2] \\
Buffer size & 20000 \\
Train start step & 200 \\
Discount Factor ($\gamma$) & 0.96 \\
Batch Size & 256 \\
Critic learning rate & 3e-4 \\
Actor learning rate & 3e-4 \\
$\alpha$ learning rate & 3e-4 \\
Target Entropy & -3 \\
Temperature ($\alpha$) & Auto tune (init 0.01) \\
Target update ratio $\tau$ & 0.005 \\
Target update frequency & 1 \\
Actor update frequency & 4 \\
Weight sync. frequency & 30 \\
\bottomrule
\end{tabular}
\end{table}

\subsubsection{SAC-Flow}

SAC-Flow extends SAC by parameterizing the policy using a continuous-time flow-based model.
The policy is defined through a velocity network $v_\theta$ that evolves a latent action variable.

SAC-Flow reparameterizes $v_\theta$  as a gated (Flow-G) or transformer-decoded (Flow-T) architecture to stabilize gradients.

\para{Deterministic Rollout.}
Given an observation $o$, a latent trajectory $\{A_{t_i}\}_{i=0}^K$ is generated via $K$ steps of Euler integration:
\begin{equation}
A_{t_{i+1}} = A_{t_i} + \Delta t_i \, v_\theta(t_i, A_{t_i}, o),
\quad A_{t_0} \sim \mathcal{N}(0, I).
\end{equation}

\para{Likelihood Construction.}
SAC requires explicit policy likelihoods for entropy regularization. Consequently, we employ a noise-augmented rollout
\begin{equation}
A_{t_{i+1}} =
A_{t_i} + v_\theta(t_i, A_{t_i}, o)\Delta t_i
+ \sigma \sqrt{\Delta t_i} \epsilon_i,
\quad \epsilon_i \sim \mathcal{N}(0, I),
\end{equation}
which preserves the marginal distribution of the final action.

\para{Actor Objective.}
The actor minimizes the joint path density $p_c(\mathcal{A}|s)$ over the $K$ sampling steps:
\begin{equation}
L_\pi(\theta) =
\mathbb{E}_{\mathcal{A} \sim \pi_\theta}
\big[
\alpha \log p_c(\mathcal{A} \mid o)
- Q_\psi(o, \tanh(A_{t_K}))
\big], 
\end{equation}
where $\mathcal{A} = \left( A_{t_0}, ..., A_{t_K} \right)$ is the intermediate action path.

We integrate SAC-Flow into our framework. The hyperparameters for the flow model, such as denoising steps and the velocity backbone, are detailed in \Cref{tab:sacflow_params}.

\begin{table}[htbp]
\centering
\caption{Hyperparameters of RLPD}
\label{tab:rlpd_params}
\begin{tabular}{ll}
\toprule
Parameter & Value \\
\midrule
Critic Network & ResNet10 encoder (shared) + MLP \\
Actor Network & ResNet10 encoder (shared) + MLP \\
Buffer size & 20000 \\
Train start step & 200 \\
Critic ensemble size & 10 \\
Critic sub-sample size & 2 \\
Demo sampling ratio & 50\% \\
Discount Factor ($\gamma$) & 0.96 \\
Batch Size & 256 \\
Critic learning rate & 3e-4 \\
Actor learning rate & 3e-4 \\
$\alpha$ learning rate & 3e-4 \\
Target Entropy & -3 \\
Temperature ($\alpha$) & Auto tune (init 0.01) \\
Target update ratio $\tau$ & 0.005 \\
Target update frequency & 1 \\
Actor update frequency & 4 \\
Weight sync. frequency & 32 \\
\bottomrule
\end{tabular}
\end{table}
\subsubsection{RLPD}
RLPD combines offline demonstration data with online exploration to significantly improve sample efficiency.

\para{Balanced Sampling.}
At each update step, mini-batches are constructed by sampling transitions from
both the online replay buffer $\mathcal{B}_{\text{online}}$ and the demonstration
buffer $\mathcal{B}_{\text{demo}}$ at a fixed ratio:

\begin{align*}
    \text{batch} = \{ (s,a,r,s')_i \sim \mathcal{B}_{online} \cup (s,a,r,s')_j \sim \mathcal{B}_{demo} \}
\end{align*}

\para{Ensemble Critics with Layer Norm:} To handle high UTD ratios, RLPD utilizes an ensemble of $M$ Q-functions with Layer Normalization to stabilize the values:
\begin{align*}
    Q_{target} = r + \gamma (\min_{j=1,\ldots,M} Q_{\bar{\psi}_j}(s', a') - \alpha \log \pi_\theta(a'|s'))
\end{align*}

RLPD leverages offline demonstration data to guide policy learning and accelerate convergence, significantly improving sample efficiency. The hyperparameters of RLPD are provided in \Cref{tab:rlpd_params}.

\begin{table}[htbp]
\centering
\caption{Hyperparameters of HG-DAgger}
\label{tab:hgd_params}
\begin{tabular}{ll}
\toprule
Parameter & Value \\
\midrule
Network & $\pi_0$ ($\sim$3B) \\
Action chunk size & 10 \\
SFT learning rate & 2.5e-5 \\
SFT decay learning rate & 2.5e-6 \\
SFT learning rate scheduler & cosine \\
SFT weight decay& 1e-10 \\
SFT batch size & 1e-5 \\
SFT epoch & 1e-5 \\
HG-DAgger learning rate & 1e-5 \\
HG-DAgger batch size & 64 \\
HG-DAgger weight decay & 1e-8 \\
Intervention sampling ratio & 50\% \\
Weight sync. frequency & 1 \\
\bottomrule
\end{tabular}
\end{table}

\subsubsection{HG-DAgger}
HG-DAgger is an interactive imitation learning algorithm designed for safe and efficient online fine-tuning. It relies on a human expert to monitor the policy and intervene when necessary.

\para{Gating Mechanism.}
At each timestep, the executed action is selected as
\begin{equation}
a_t =
\begin{cases}
a_t^{\text{human}}, & \text{if human intervention is active}, \\
\pi_\theta(o_t), & \text{otherwise}.
\end{cases}
\end{equation}

\para{Behavior Cloning Objective.}
State-action pairs collected during the human intervention are stored in
$\mathcal{D}_{\text{intervene}}$, and the policy is updated via  a behavior cloning (BC) loss:
\begin{equation}
\mathcal{L}_{BC}(\theta) =
\mathbb{E}_{(o, a^h) \sim
\mathcal{D}_{\text{intervene}} \cup \mathcal{D}_{\text{demo}}}
\big[
\| \pi_\theta(o) - a^h \|^2
\big].
\end{equation}

We utilize HG-DAgger to fine-tune $\pi_0$ via human interventions. We first perform Supervised Fine-Tuning (SFT) on a pre-collected dataset to initialize the $\pi_0$ model, providing the policy with a non-zero initial success rate. The parameters of SFT and HG-DAgger are listed in Table \ref{tab:hgd_params}.

\subsection{Task Implementation Details}
\label{app:task_details}
In this section, we detail task settings and task-specific training configurations. \Cref{fig:tasks} shows all five tasks used for evaluation.

\subsubsection{Peg Insertion}
The task requires the robot to perform high-precision insertion of a peg into its corresponding slot. The static pose of the slot enables straightforward success verification and automated resetting, making this task ideal for real-world RL.

Table \ref{tab:peg_params} reports the task-specific training settings for the Peg Insertion task. In this task, we use a single wrist camera as the visual observation. The state observation includes the pose, velocity, force, and torque of the end-effector (EE), and the gripper pose. The reward signal is derived directly from the distance between the end-effector and the target pose. 

\begin{table}[htbp]
\centering
\caption{Detailed settings of peg insertion}
\label{tab:peg_params}
\begin{tabular}{ll}
\toprule
Parameter & Value \\
\midrule
Image shape & 1$\times$ (3, 128, 128) \\
State dim & 19\\
Obs coordinate & Reset pose \\
Action space & Delta EE pose (6 dim) \\
Step frequency & 10 Hz \\
Max episode length & 100 steps \\
Bounding box (xyz) & [0.05, 0.05, 0.1] (m) \\
Bounding box (rpy) & $\pi$/6 (rad) \\
\midrule
\textbf{RLPD Specific:} & \\
Demos & 20 \\
Demo Sampling Ratio & 50\% \\
Reward & Rule-based, sparse \\
\midrule
\textbf{SAC Specific:} & \\
Demos & / \\
Reward & Rule-based, dense \\
\midrule
\textbf{SAC Flow Specific:} & \\
Demos & 20 \\
Reward & Rule-based, sparse \\
\bottomrule
\end{tabular}
\end{table}

\subsubsection{Charger Plugging}
The charger plugging task involves inserting a charger into a socket. This task presents a significant challenge in high-precision manipulation, defined by the narrow tolerance of the socket geometry, the sparsity of visual features, and the visual occlusion during the approach phase. To address these challenges, we employ a desktop third-view camera for visual feedback.

\Cref{tab:charger_params} details task configurations and task-specific training settings. Given that the task necessitates sub-millimeter precision, we utilized a more constrained bounding box to focus the operational workspace.

\begin{table}[htbp]
\centering
\caption{Detailed settings of charger plugging}
\label{tab:charger_params}
\begin{tabular}{ll}
\toprule
\textbf{Parameter} & \textbf{Value} \\
\midrule
Image shape & 1$\times$ (3, 128, 128) \\
State dim & 19\\
Obs coordinate & Reset pose \\
Action space & Delta EE pose (6 dim) \\
Step frequency & 10 Hz \\
Max episode length & 100 \\
Bounding box (xyz) & [0.02, 0.02, 0.055] (m) \\
Bounding box (rpy) & $\pi$/9 (rad) \\
\midrule
\textbf{RLPD Specific:} & \\
Demos & 20 \\
Demo Sampling Ratio & 50\% \\
Reward & Rule-based, sparse \\
\midrule
\textbf{SAC Specific:} & \\
Demos & / \\
Reward & Rule-based, dense \\
\midrule
\textbf{SAC Flow Specific:} & \\
Demos & 20 \\
Reward & Rule-based, sparse \\

\bottomrule
\end{tabular}
\end{table}

\subsubsection{Cap Tightening}
The cap tightening task requires the policy to drive a pre-positioned cap toward a target configuration. For the required angular displacement that spans multiple full rotations, the policy must learn a cyclic regrasping strategy to facilitate continuous, multi-turn manipulation.

We equip a wrist camera and a stationary third-view camera. This task requires higher rotation relative to translational displacement, so we employ a specific bounding box. Reward signals are sparsely provided by a human operator, who utilizes a foot-pedal interface to assign a binary success signal upon completion. Environment resets are managed manually as well. The detailed task settings are available in \Cref{tab:cap_params}.

\begin{table}[htbp]
\centering
\caption{Detailed settings of cap tightening}
\label{tab:cap_params}
\begin{tabular}{ll}
\toprule
\textbf{Parameter} & \textbf{Value} \\
\midrule
Image shape & 2$\times$ (3, 128, 128) \\
State dim & 20\\
Obs coordinate & Target pose \\
Action space & Delta EE pose + gripper (7 dim) \\
Step frequency & 10 Hz \\
Max episode length & 240 \\
Bounding box (xyz) & [0.01, 0.01, 0.02] (m) \\
Bounding box (rpy) & $\pi$/6 (rad)\\
\midrule
\textbf{RLPD Specific:} & \\
Demos & 20 \\
Demo Sampling Ratio & 50\% \\
Reward & Human-provided, sparse \\
\bottomrule
\end{tabular}
\end{table}

\subsubsection{Pick-and-Place}
The pick-and-place task requires the policy to successfully transfer objects between two trays. This task is characterized by a vast exploration space, posing significant challenges for sample efficiency. We employ a wrist camera and a third-view camera at a 45-degree downward view. This combination ensures the policy has access to both fine-grained contact information and broad spatial context. Following our human-in-the-loop framework, both reward signals and environment resets are mediated by a human operator, as detailed in \Cref{tab:pnp_params}.

\begin{table}[htbp]
\centering
\caption{Detailed settings of Pick-and-Place}
\label{tab:pnp_params}
\begin{tabular}{ll}
\toprule
\textbf{Parameter} & \textbf{Value} \\
\midrule
Image shape & 2$\times$ (3, 128, 128) \\
State dim & 20\\
Obs coordinate & Reset pose \\
Action space & Delta EE pose + gripper (7 dim) \\
Step frequency & 10 Hz \\
Max episode length & 240 \\
Bounding box (xyz) & [0.3, 0.3, 0.15] (m) \\
Bounding box (rpy) & $\pi$/6 (rad) \\
\midrule
\textbf{RLPD Specific:} & \\
Demos & 40 \\
Demo Sampling Ratio & 50\% \\
Reward & Human-provided, sparse \\
\midrule
\textbf{HG-DAgger Specific:} & \\
Demos & 40 \\
SFT epoch & 5000 \\
\bottomrule
\end{tabular}
\end{table}

\subsubsection{Table Clean-up}
The table clean-up task is a long-horizon, multi-stage manipulation task that requires a high degree of precision. The policy must execute a specific sequence: first, placing a marker into a container (orange box); second, nesting that container within a larger receptacle (white box); and finally, closing a semi-transparent lid. Both the grasping of the orange container and the final lid-closure demand fluid, high-precision control and contact-rich interactions. This task also integrates a wrist camera for localized, and a third-view camera to provide global spatial context.

\begin{table}[htbp]
\centering
\caption{Policy training details for Table Clean-up (HG-DAgger).}
\label{tab:cleanup_params}
\begin{tabular}{ll}
\toprule
Parameter & Value \\
\midrule
Image shape & 2$\times$ (3, 128, 128) \\
State dim & 20\\
Obs coordinate & Reset pose \\
Action space & Delta EE pose + gripper (7 dim) \\
Step frequency & 10 Hz \\
Max episode length & 360 \\
Bounding box (xyz) & [0.4, 0.4, 0.25] (m) \\
Bounding box (rpy) & $\pi$/3 \\
\midrule
\textbf{HG-DAgger Specific:} & \\
Demos & 40 \\
SFT epoch & 15000 \\
\bottomrule
\end{tabular}
\end{table}

\subsection{Comparison between USER and other real-world learning systems}
We compare the convergence performance of USER with other learning systems on the peg-insertion and charger tasks. Since SOP and Qt-Opt are closed-source, and other frameworks (e.g., RLlib, TMRL) lack native real-world support, we benchmark USER against SERL. As shown in \Cref{fig:serl_compare}, USER achieves training speeds comparable to or exceeding SERL across both tasks.

\begin{figure}[hbp]
    \centering
    \subfloat[Peg Insertion]{
        \includegraphics[width = 0.7\linewidth]{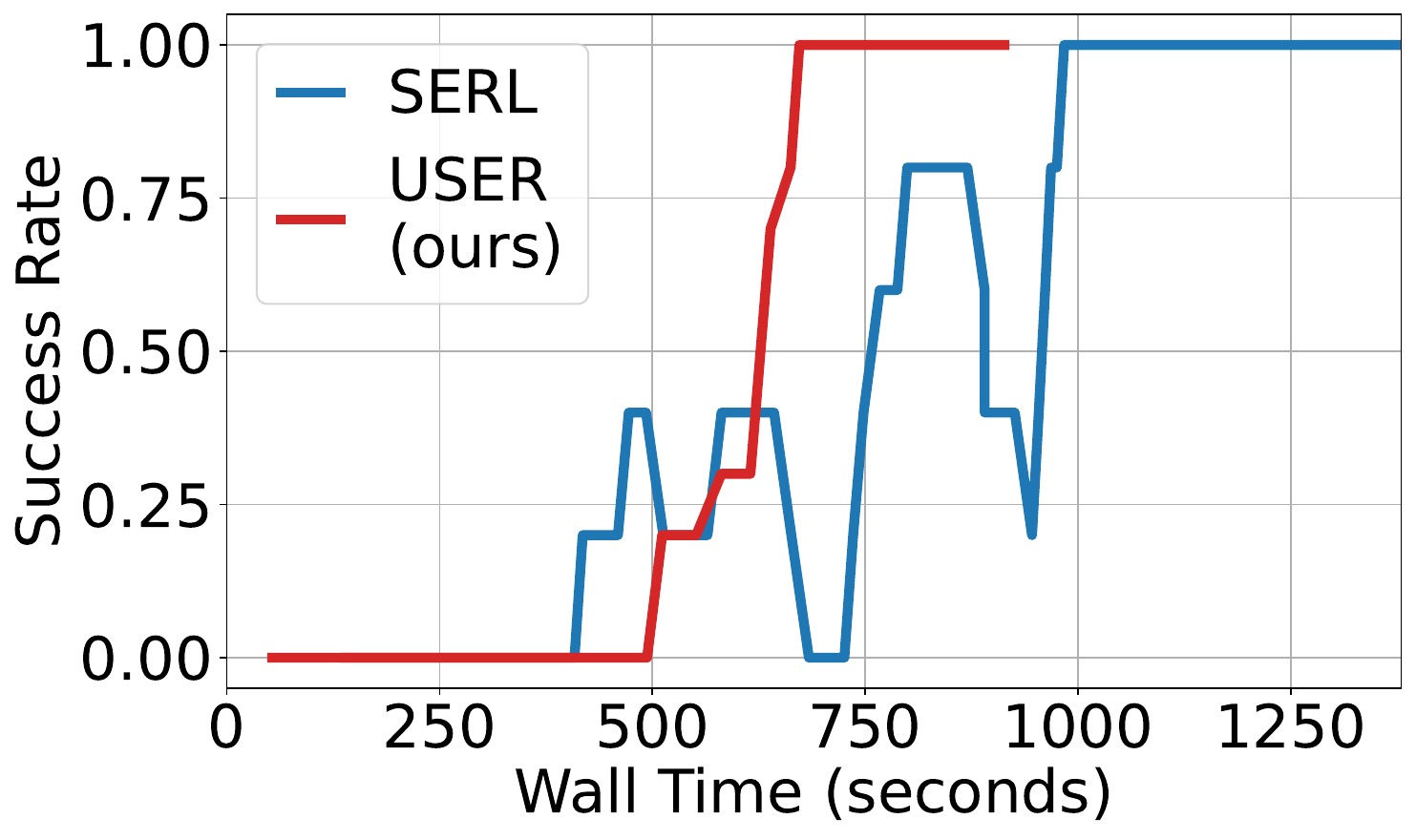}
    }
    \\
    \subfloat[Charger]{
        \includegraphics[width = 0.7\linewidth]{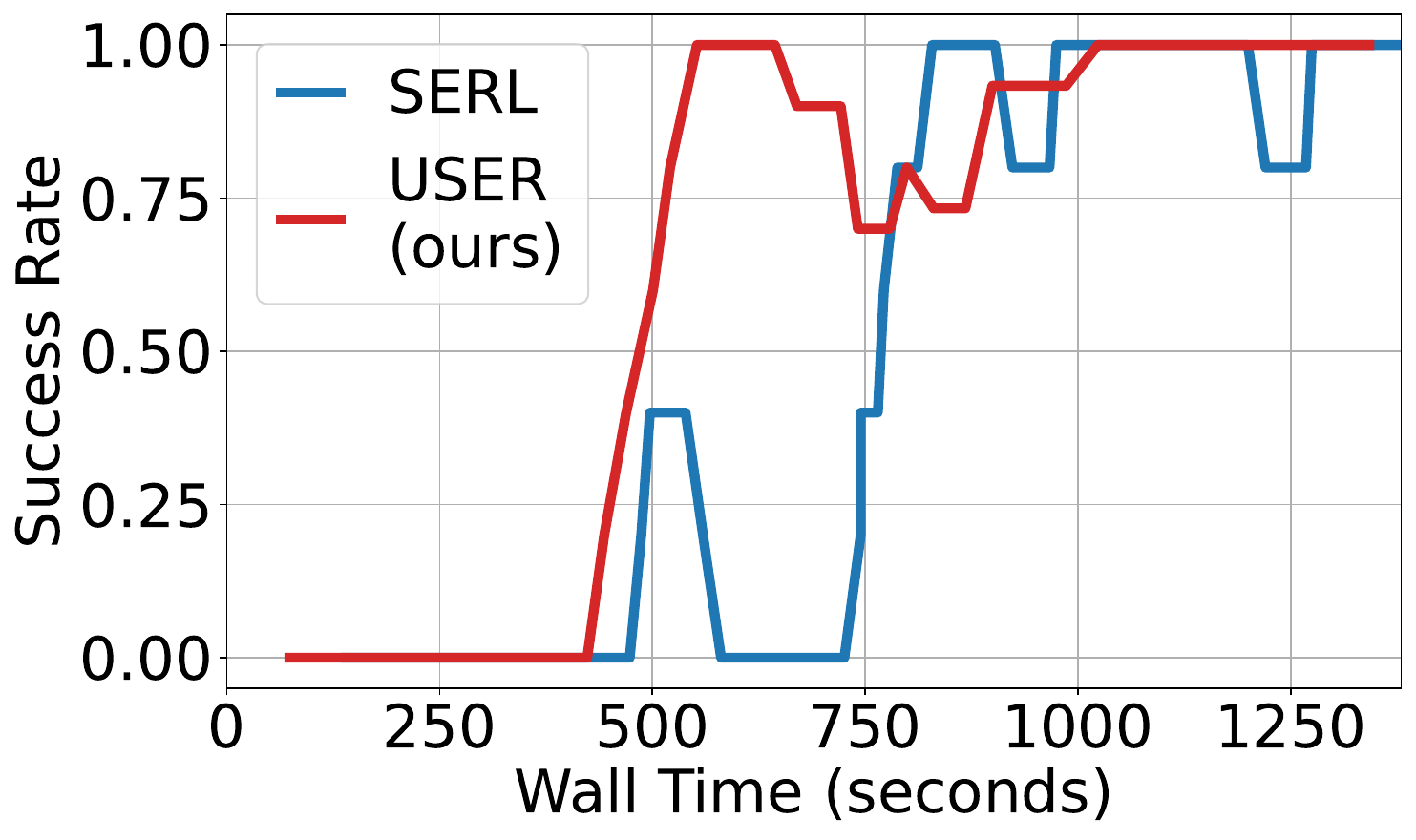}
    }
    \caption{USER vs. SERL: Single-robot convergence comparison}
    \label{fig:serl_compare}
\end{figure}

\subsection{Hardware Details}
\label{app:hardware_setup}
We conduct all experiments across two distinct robotic platforms: a 7-DoF Franka robot arm and a 6-DoF low-cost ARX robot arm (\Cref{fig:robots}). This section provides a detailed description of the robotic hardware configurations.

\begin{figure}
  \centering 
  \subfloat[Franka arm]{%
    \includegraphics[height=1.7in]{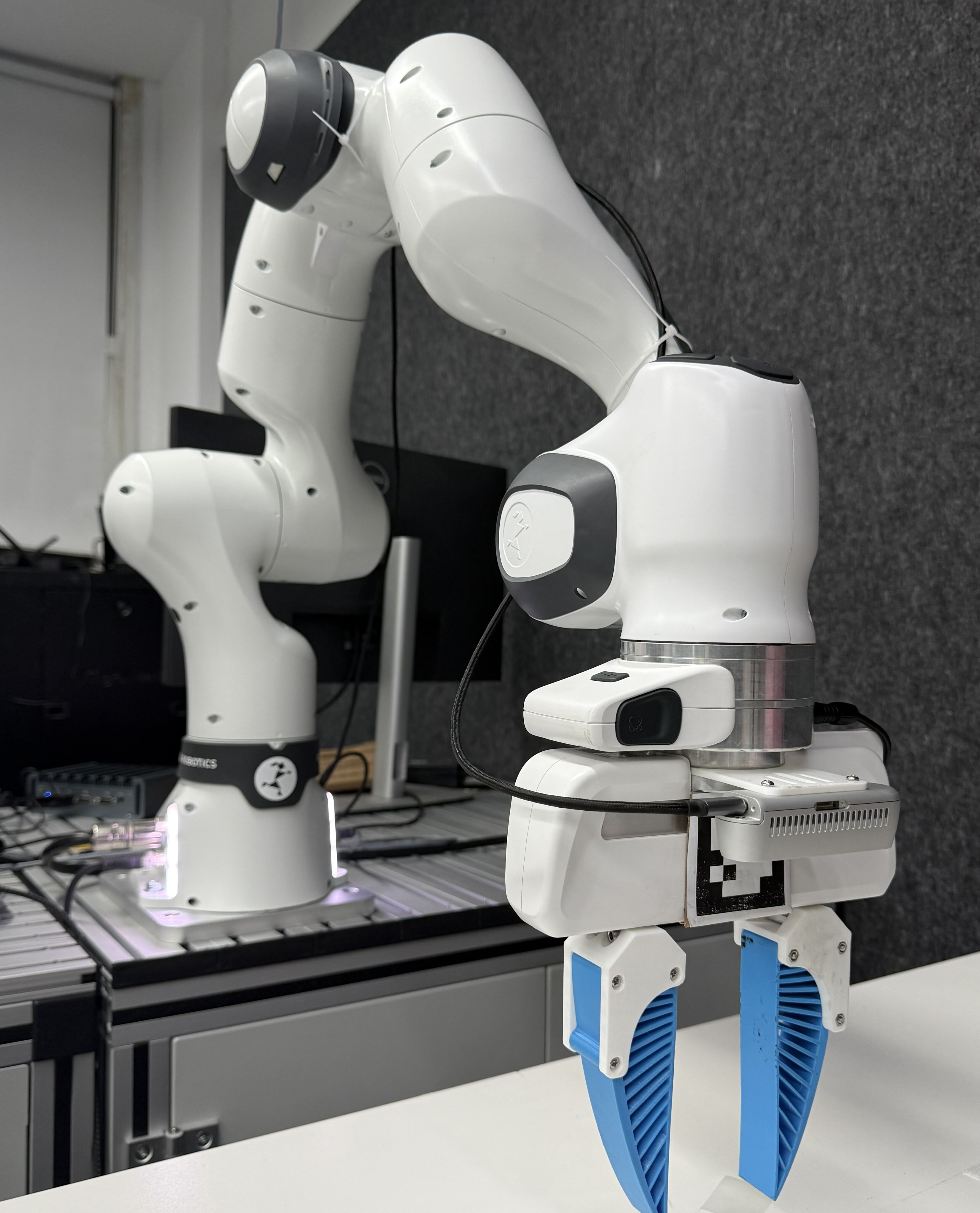}%
    \label{fig:robot-franka} 
  }
  \subfloat[ARX arm]{%
    \includegraphics[height=1.7in]{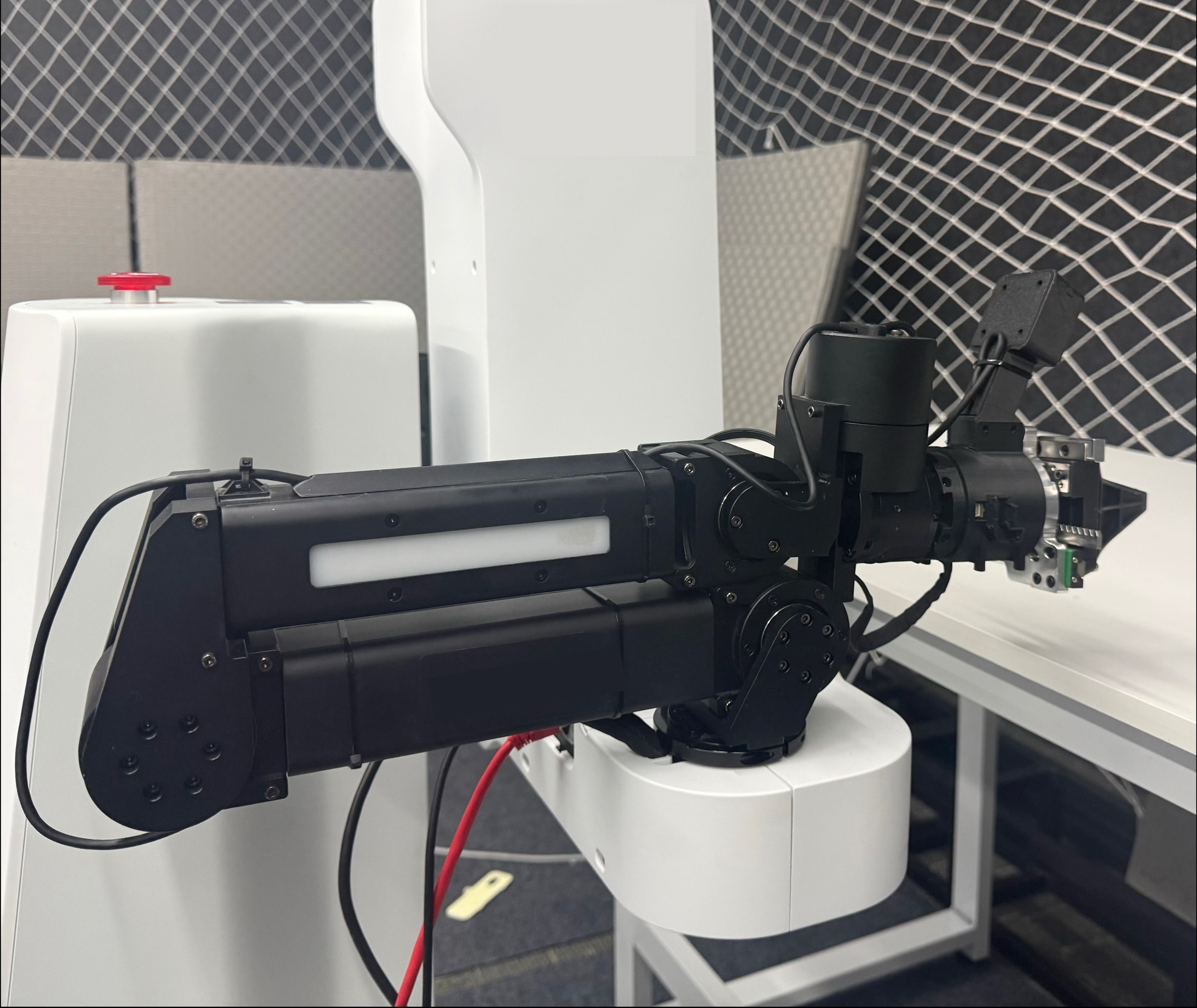}%
    \label{fig:robot-arx} 
  }
  
  \caption{We run our experiments on two types of robot arms. (a) 7-DoF Franka robot arm, (b) Low-cost 6-DoF ARX robot arm.}
  \label{fig:robots}
\end{figure}

\subsubsection{Franka Robot Arm}
Most of our experiments were conducted using a Franka Emika Panda research arm. The Franka Emika Panda is a 7-DOF collaborative robot arm, widely utilized in robotics research for its high execution precision and integrated sensing capabilities. The robot arm features highly sensitive torque sensors in all seven joints, providing force and torque inputs for the policy. We use Franka Control Interface (FCI) to control the robot via libfranka. We utilize external RealSense D435i cameras to provide perception.

RL involves interacting with the environment through trial and error. Consequently, we employ an impedance controller following SERL~\cite{luo2024serl}. Compared to standard position controllers, impedance control offers significant advantages in contact-rich tasks. By regulating the dynamic relationship between interaction forces and motion, the impedance controller effectively models the robot as a tunable mass-spring-damper system. It allows the policy to exhibit inherent compliance during collisions rather than rigidly tracking trajectories, ensuring hardware safety during exploration.

Our RL policy sends control targets at 10 Hz for the low-level impedance controller to track at 1kHz. A typical impedance control objective is
\[
    F = k_p \cdot e + k_d \cdot \dot{e} + F_{ff} + F_{cor},
\]
where $e = p-p_{ref}$, $p$ is the measured pose, and $p_{ref}$ is the target pose. $F_{ff}$ is the feed-forward force, $F_{cor}$ is the Coriolis force. To map this objective into the joint space, we compute the required torques using the Jacobian transpose. The resulting control law emulates a PD-controlled spring-damper system, where the system tracks the equilibrium $p_{ref}$ with stiffness and damping gains defined by $k_p$ and $k_d$, respectively. The parameters of this impedance controller are detailed in \Cref{tab:compliance_params}.

\begin{table}[htbp]
\centering
\caption{Parameters of the impedance controller}
\label{tab:compliance_params}
\begin{tabular}{ll|ll}
\toprule
\textbf{\makecell[l]{Translational\\Parameter}} & \textbf{Value} & \textbf{\makecell[l]{Rotational\\Parameter}} & \textbf{Value} \\
\midrule
Stiffness & 2500 & Stiffness & 150 \\
Damping & 100 & Damping & 7 \\
$K_i$ & 0 & $K_i$ & 0 \\
Clip x & [-0.007, 0.007] & Clip x & [-0.07, 0.07] \\
Clip y & [-0.007, 0.007] & Clip y & [-0.07, 0.07] \\
Clip z & [-0.007, 0.007] & Clip z & [-0.06, 0.06] \\
\bottomrule
\end{tabular}
\end{table}

\subsubsection{ARX arm}
The ARX-R5 is a cost-effective, 6-DOF robot arm. Due to the absence of force/torque sensors, we employ a simple PD position controller for operation. The RL policy sends position targets at 10 Hz, and the position controller tracks at 200 Hz. The manipulator is equipped with a wrist fisheye camera, as illustrated in \Cref{fig:robot-arx}.

\fi

\end{document}